%% file: main.tex
\documentclass{article}

\usepackage{microtype}
\usepackage{graphicx}
\usepackage{subcaption}
\usepackage{booktabs} 

\usepackage[preprint]{icml2026}

\usepackage{amsmath}
\usepackage{header}
\usepackage{amssymb}
\usepackage{mathtools}
\usepackage{amsthm}
\input{iclr2025/math_commands}

\usepackage[textsize=tiny]{todonotes}

\icmltitlerunning{Reinforcement Learning with Discrete Diffusion Policies for Combinatorial Action Spaces}

\begin{document}

\twocolumn[
  \icmltitle{Reinforcement Learning with Discrete Diffusion Policies\\ for Combinatorial Action Spaces}



  \icmlsetsymbol{equal}{*}
  \icmlsetsymbol{equal_sup}{\dag}

\begin{icmlauthorlist}
\icmlauthor{Haitong Ma}{equal,google_res,harvard}
\icmlauthor{Ofir Nabati}{equal,google_res,technion}
\icmlauthor{Aviv Rosenberg}{google_res}
\icmlauthor{Bo Dai}{deepmind}
\icmlauthor{Oran Lang}{google_res}
\icmlauthor{Craig Boutilier}{google_res}
\icmlauthor{Na Li}{harvard}
\icmlauthor{Shie Mannor}{technion,nvidia}
\icmlauthor{Lior Shani}{equal_sup,google_res}
\icmlauthor{Guy Tenneholtz}{equal_sup,google_res}
\end{icmlauthorlist}

\icmlaffiliation{google_res}{Google Research}
\icmlaffiliation{harvard}{Harvard University}
\icmlaffiliation{technion}{Technion}
\icmlaffiliation{deepmind}{Google DeepMind}
\icmlaffiliation{nvidia}{Nvidia Research}

\icmlcorrespondingauthor{Haitong Ma and Ofir Nabati}{\{haitongma8,ofirnabati\}@gmail.com}

  \icmlkeywords{Machine Learning, ICML}

  \vskip 0.3in
]



\printAffiliationsAndNotice{\icmlEqualContribution\icmlEqualSupervision}  

\begin{abstract}
  Reinforcement learning (RL) struggles to scale to large, combinatorial action spaces common in many real-world problems. This paper introduces a novel framework for training discrete diffusion models as highly effective policies in these complex settings. 
    Our key innovation is an efficient online training process that ensures stable and effective policy improvement. By leveraging policy mirror descent (PMD) to define an ideal, regularized target policy distribution, we frame the policy update as a distributional matching problem, training the expressive diffusion model to replicate this stable target. This decoupled approach stabilizes learning and significantly enhances training performance.
    Our method achieves state-of-the-art results and superior sample efficiency across a diverse set of challenging combinatorial benchmarks, including DNA sequence generation, RL with macro-actions, and multi-agent systems. Experiments demonstrate that our diffusion policies attain performance comparable or superior to baseline methods. Crucially, our extensive empirical analysis reveals a key trade-off: FKL demonstrates superior sample efficiency and faster initial convergence, whereas RKL ensures stable training and higher asymptotic performance on challenging tasks. 
\end{abstract}

\input{contents/1_intro}
\input{contents/2_related}
\input{contents/3_problem}
\input{contents/4_algorithm}

\input{contents/5_experiments}
\input{contents/6_conclusion}

\bibliography{icml2026/icml2026}
\bibliographystyle{icml2026}

\newpage
\appendix
\onecolumn
\input{contents/A_appendix}

\end{document}

%% file: iclr2025/math_commands.tex
\usepackage{amsmath,amsfonts,bm}

\def\eqref#1{equation~\ref{#1}}

\def\1{\bm{1}}

\DeclareMathAlphabet{\mathsfit}{\encodingdefault}{\sfdefault}{m}{sl}
\SetMathAlphabet{\mathsfit}{bold}{\encodingdefault}{\sfdefault}{bx}{n}

\def\gA{{\mathcal{A}}}

\def\gD{{\mathcal{D}}}

\def\gH{{\mathcal{H}}}

\def\gL{{\mathcal{L}}}

\def\gS{{\mathcal{S}}}

\def\sR{{\mathbb{R}}}

\DeclareMathOperator*{\argmax}{arg\,max}
\DeclareMathOperator*{\argmin}{arg\,min}

%% file: contents/1_intro.tex
\section{Introduction}

Reinforcement learning (RL) has been instrumental in pushing the boundaries of autonomous decision-making, achieving superhuman performance in a diverse range of complex sequential tasks \cite{silver2016mastering, vinyals2019grandmaster, schrittwieser2020mastering}. However, a significant frontier remains: scaling these successes to problems with vast, combinatorial discrete action spaces. Such challenges are not niche; they are central to many real-world applications, including planning with macro-actions in hierarchical RL \cite{sutton1999between, durugkar2016deep}, coordinating strategies in multi-agent systems \cite{hernandez2019survey}, and generating slates in recommender systems~\cite{slateQ:ijcai19}. The sheer scale of these action spaces poses a fundamental challenge to standard RL algorithms, demanding highly efficient policy parameterizations and effective exploration strategies.

Prior approaches have attempted to mitigate this complexity by mapping actions to lower-dimensional subspaces \cite{stulp2012reinforcement,tennenholtz2019natural}, employing hierarchical training schemes \cite{nachum2018data}, or assuming specific structural properties of the action space \cite{carrara2019budgeted}. While effective in certain contexts, these methods often rely on structural assumptions or inductive biases that may not hold in more general and complex problem settings. More recently, the success of autoregressive models \cite{vaswani2017attention} has inspired their use for policies over combinatorial actions \cite{chen2021decision, wen2022multi}. Yet, these models suffer from two key limitations: high computational cost during inference due to their sequential generation process and the imposition of a causal action ordering, which is often an artificial and restrictive constraint.

Diffusion models have emerged as a powerful class of generative models, renowned for their ability to capture highly complex probability distributions without imposing a causal structure \cite{sohl2015deep, ho2020denoising}. Recent extensions to discrete spaces have further broadened their applicability \cite{austin2021structured, sun2022score, campbell2022continuous, shi2024simplified}. This inherent flexibility and expressiveness make them an ideal candidate for modeling policies in large, unstructured discrete action spaces. While diffusion models have been actively explored for synthesizing policies in continuous control \cite{wang2022diffusion, ding2024diffusion, ren2024diffusion, ma2025soft}, a principled and efficient framework for training discrete diffusion policies with RL remains unexplored.

In this work, we introduce a novel framework for training discrete diffusion models as highly effective policies for combinatorial action spaces. Our key innovation is an efficient online training process that ensures stable and effective policy improvement. We leverage policy mirror descent (PMD, \citet{shani2020adaptive,tomar2021mirrordescentpolicyoptimization, lan2023policy}) to define the ideal target policy distribution based on the PMD optimization objective. This reframes the policy update as a distributional matching problem, where we train our expressive diffusion model to replicate this stable target. This decoupled approach is critical: it separates the RL objective optimization from the complex task of representation learning, which we delegate to the diffusion model, thereby stabilizing the entire learning process and significantly enhancing performance.

Our core contributions are as follows: (1) We introduce RL-D$^2$, a new and efficient online training framework for using discrete diffusion models as policies in RL for combinatorial action spaces. Our core mechanism reframes the policy update as a distributional matching problem by using policy mirror descent (PMD) to define a stable target distribution, which significantly stabilizes learning. (2) We derive and analyze two practical policy improvement methods based on minimizing the forward and reverse Kullback-Leibler (KL) divergence to the PMD target. (3) Finally, we conduct extensive experiments across three distinct and challenging domains: DNA sequence generation~\cite{gosai2023machine}, long-horizon RL with macro-actions in Atari~\cite{bellemare2013arcade}, and cooperative multi-agent RL in the challenging Google Research Football domain~\cite{kurach2020google}. In all settings, our method achieves state-of-the-art results, demonstrating superior performance, scalability, and efficiency.


    

%% file: contents/2_related.tex
\section{Related Work}




\textbf{Discrete Diffusion Models.} Diffusion models for generating continuous data, such as images, typically rely on the gradual addition and removal of Gaussian noise to learn and synthesize complex probability distributions~\citep{sohl-dickstein2015deep,ho2020denoising}. However, this paradigm is ill-suited for discrete data like text or biological sequences, where values are categorical and adding small amounts of continuous noise is not meaningful. To address this, \emph{discrete diffusion models} extend the iterative refinement idea to discrete state spaces, with forward and backward processes using Markov chains where each transition in a sequence is sampled independently. Various approaches have explored different transition mechanisms and training objectives~\citep{austin2021structured,campbell2022continuous,sun2022score,lou2023discrete,shi2024simplified}.
Among these, \emph{absorbing (or masked) diffusion} has proven to be particularly effective~\citep{sahoo2024simple,ou2024your}. The success of these models has led to their application in a range of domains. In natural language processing, they have been adapted for complex generation tasks~\citep{arriola2025block,ye2025dream,nie2025large}. More relevant to our work, discrete diffusion has shown significant promise in bio-sequence modeling for generating novel proteins and DNA sequences with desired properties~\citep{gruver2023protein,wang2024fine}.

\textbf{Reward-based Fine-tuning and RL for Discrete Diffusion.} A key challenge, beyond unconditional generation, is adapting discrete diffusion to optimize for specific objectives. This has primarily been approached through reward-based fine-tuning, which adjusts the model's parameters to increase the likelihood of generating high-reward samples. For instance, \cite{wang2024fine} enable direct reward backpropagation by leveraging the Gumbel-softmax trick, while \cite{zekri2025fine} optimize the model by manipulating the score entropy~\citep{lou2023discrete}. While effective, these methods can be viewed as forms of single-step policy optimization. By contrast, the application of online RL to discrete diffusion remains unexplored and faces challenges such as exploration-exploitation trade-offs, computational efficiency, and horizon-complexity trade-offs.

Although black-box methods like Evolution Strategies \citep{salimans2017evolution, conti2018improving, nabati2023representation}, Genetic Algorithms \citep{such2017deep}, and the Cross-Entropy Method \citep{de2005tutorial} offer alternative training paradigms, our approach provides two key advantages over standard discrete solvers. First, instead of optimizing static, single-step objectives, we leverage multi-step transition dynamics, using the discrete diffusion policy as an expressive prior to efficiently guide exploration in large combinatorial spaces. Second, our FKL variant supports off-policy training, maximizing sample efficiency through data reuse.



%% file: contents/3_problem.tex
\section{Preliminaries}
\label{sec:preliminaries}

In this section, we review the necessary background. We first define the problem setup for RL with large, combinatorial action spaces. We then introduce policy mirror descent as the foundation for our policy improvement step, followed by a review of discrete diffusion models, which will serve as our policy parameterization.

\subsection{Problem Setup}
\label{sec.problem_formulation}
We consider a \emph{Markov decision process (MDP)} defined by a tuple $(\gS, \gA^K, P, \gamma, r, \rho_0)$, where $\gS$ is the state space, $P$ is the transition function, $r$ is a reward function, $\gamma \in [0, 1)$ is the discount factor, and $\rho_0$ is the initial state distribution. The action space $\gA^K$ is assumed to be large, with some combinatorial structure, i.e., $\textbf{a} \in \gA^K$ is a structured, ``multi-component'' object\footnote{For simplicity we focus on power sets of $\gA$, though more complex combinatorial action spaces can be used.}. The reward function $r: \gS \times \gA^K \to \sR$ and transition function $P: \gS \times \gA^K \mapsto \Delta_{\gS}$ are defined w.r.t.\ $\gA^K$.

This general setup encapsulates a range of different problems, one of which is \emph{hierarchical RL} \citep{sutton1999between, vezhnevets2017feudal, haarnoja2018latent}, where each action $\textbf{a} \in \gA^K$ is a \emph{macro-action}, or a sequence of $K$ primitive actions, $\textbf{a} = (a_1, \hdots, a_{K})$.\footnote{We abuse terminology slightly. In general, macro-actions (or \emph{options}) are general ``local'' policies with suitable termination conditions that can be used within a larger hierarchical or abstract policy \citep{Sutton-etal:AIJ99,HMKDB:uai98}. However, the fixed sequence view of macros (a special case of the former) also appears in the literature \citep{durugkar2016deep}.}. In this case, $r(s, \textbf{a})$ and $P(s' | s, \textbf{a})$ represent the total discounted reward and the final state after executing the entire $K$-step sequence. Other examples include multi-agent policy optimization \citep{hernandez2019survey}, where $\textbf{a} = (a_1, \dots, a_K)$ is the joint action for $K$ agents and $a_i\in \gA_i$ is the $i$th agent's action; 
slate recommendation \citep{slateQ:ijcai19}, where $a_i$ is the item at the $i$th position of a set/slate of recommendations of size $K$; and combinatorial sequential assignment \citep{carrara2019budgeted}.

A policy $\pi: \gS \mapsto \Delta_{\gA^K}$ maps a state to a distribution over the action space. The state-action value function $q^{\pi}(s, \textbf{a})$ is the expected return after taking action $\textbf{a}$ in state $s$ and following $\pi$ thereafter:
\begin{equation}
q^{\pi}(s, \textbf{a}) = r(s, \textbf{a}) + \gamma \expect*{s' \sim P(\cdot|s, \textbf{a}), \textbf{a}' \sim \pi(\cdot|s')}{q^\pi(s', \textbf{a}')}.
\label{eq:q_def}
\end{equation}
The state-value function is the expectation over actions, $v^\pi(s) := \expect*{\textbf{a} \sim \pi(\cdot|s)}{q^\pi(s,\textbf{a})}$. The agent's goal is to find an optimal policy $\pi^*$ within a policy class $\Pi$ that maximizes the expected return: $\pi^* \in \argmax_{\pi \in \Pi} \expect*{s \sim \rho_0}{v^\pi(s)}$.

\begin{remark}[Compared to discrete black-box solver.]
Compared with standard discrete black-box solvers, our RL-D$^2$ framework is particularly well suited to sequential decision-making problems with large combinatorial action spaces. While methods such as evolutionary algorithms and zeroth-order optimization typically treat each problem instance as a static objective and often optimize from scratch, RL-$D^2$ learns a state-conditioned diffusion policy that can exploit transition dynamics, perform structured exploration, and generalize to unseen states without re-solving each instance independently.    
\end{remark}

\subsection{Policy Mirror Descent}
\label{sec:pmd}
\emph{Policy mirror descent (PMD)}~\citep{beck2003mirror,shani2020adaptive,tomar2020mirror, lan2023policy} is a policy optimization method that provides a provably convergent, stable, and regularized policy improvement step. Given a current policy $\pi_{\rm old}$, the PMD update finds a new policy $\pi$ by solving:
\begin{equation}
\pi(\cdot |s) \in \argmax_{\pi \in \Pi} \expect*{\textbf{a} \sim \pi(\cdot|s)}{A^{\pi_{\rm old}}(s, \textbf{a})} - \lambda d_{KL}(\pi,\pi_{\rm old};s)
\label{eq:general_pmd}
\end{equation}
$\forall s \in \gS$, where $A^{\pi_{\rm old}}(s,\textbf{a}) := q^{\pi_{\rm old}}(s,\textbf{a}) - v^{\pi_{\rm old}}(s)$ is the advantage function, $\lambda > 0$ is a temperature parameter, and $d_{KL}$ is the Kullback-Leibler (KL) divergence:
$d_{KL}(\pi, \mu ;s) := \sum_{\textbf{a} \in \gA^K} \pi(\textbf{a}|s) \log \frac{\pi(\textbf{a}|s)}{\mu(\textbf{a}|s)}$.

The unique solution to this optimization problem is given by:
\begin{equation}
\pi_{\rm MD}(\textbf{a}|s) =
\pi_{\rm old}(\textbf{a}|s)\exp\brk{A^{\pi_{\rm old}}(s, \textbf{a}) / \lambda} \;/\; Z(s),
\label{eq:opt_pi}
\end{equation}
where $Z(s) = \expect{\textbf{a} \sim \pi_{\rm old}}{\exp\brk{A^{\pi_{\rm old}}(s, \textbf{a}) / \lambda}}$ is the normalization constant, or partition function.

\subsection{Masked Discrete Diffusion Processes}
\label{sec:diffusion_prelim}
We provide a general background on discrete diffusion in this subsection, and refer the reader to \citet{austin2021structured} and~\Cref{sec.appendix.diffusion} for an exhaustive derivation of this method. A reader already familiar with discrete diffusion processes can skip directly to \Cref{sec:method}.

Discrete diffusion models are powerful generative models, well-suited for capturing complex distributions over structured, sequential data. We therefore focus on combinatorial action spaces that can be represented as a fixed-length sequence of $K$ discrete actions, $\textbf{a} = (a_0, \dots, a_{K-1}) \in \gA^K$. This formulation directly applies to the macro-action problem and can be adapted for other settings like multi-agent joint actions by imposing a consistent ordering on the agents. We use a masked diffusion process \citep{shi2024simplified}, which operates over an augmented vocabulary $\gA \cup \{\texttt{m}\}$ that includes a mask action $m$.

\paragraph{Forward Process.}
The fixed forward process $q$ gradually noises a clean macro-action $\textbf{a}^0 \in \gA^K$ into a fully masked sequence $\textbf{a}^N$ over $N$ discrete steps. This noising process is applied independently to each component $a_t \in \textbf{a}$. The single-action transition is defined as
$$
q\left(a^n \mid a^{n-1}\right) = \begin{cases}
    \beta_n & \text{if } a^{n-1}\neq\texttt{m} \text{ and } a^n=\texttt{m}\\
    1-\beta_n  & \text{if } a^{n-1}\neq\texttt{m} \text{ and } a^n=a^{n-1}\\
    1 & \text{if } a^{n-1}=a^n=\texttt{m}\\
    0 &\text{otherwise}
\end{cases},
$$
 where $\{\beta_n\}_{n=1}^N$ is a fixed schedule. This defines a marginal distribution 
 $$
 q(a^n | a^0) = \text{Cat}\left(a^{n}; \alpha_n a^0 + (1 - \alpha_n)\textbf{e}_m \right),
 $$ 
 where $a^n = a^0$ with probability $\alpha_n$ and $a^n = m$ with probability $1 - \alpha_n$, for a known noise schedule $\alpha_n$.

\paragraph{Reverse Process.}
The learned reverse process $p_\theta(\textbf{a}^{n-1} | \textbf{a}^n, s)$ is trained to reverse the forward process, conditioned on the state $s$. It iteratively denoises a sequence $\textbf{a}^n$, starting from the pure noise prior $\textbf{a}^N \sim p(\cdot|s)$, to generate a clean macro-action $\textbf{a}^0 \sim \pi_\theta(\cdot|s)$. This process is parameterized by the posterior:
\begin{align*}
    &q(a^{n-1} | a^n, a^0) \\ &= \begin{cases} \text{Cat}\left(a^{n-1}; \bar \alpha_n a^0 + (1 - \bar \alpha_n)\textbf{e}_m \right) & \quad a^n = \textbf{e}_m \\
\text{Cat}\left(a^{n-1}; a^n \right) & \quad a^n \neq \textbf{e}_m
\end{cases}
\end{align*}

where $\bar \alpha_n := \frac{\alpha_{n-1} - \alpha_n}{1 - \alpha_n}$, through the parameterized model $p_\theta(a^{n-1} | a^n) := q(a^{n-1}|a^n,  \mu_\theta(a^n,n))$, where 
\[
\mu_\theta(a^n, n) = \begin{cases}
\text{softmax}(f_\theta(a^n, n)) & \quad a^n = \texttt{m}, \\
a^n & \quad a^n \neq \texttt{m}.
\end{cases}
\]
Here, $\mu_\theta$ is the clean sample mean-value estimator induced by a trained model $f_\theta$ (e.g., a Transformer) that predicts the clean sequence $\mu_\theta(\textbf{a}^n, n, s) \approx \textbf{a}^0$ from any noised sequence $\textbf{a}^n$ at step $n$.

\paragraph{Training Objective.}
The model $f_\theta$ is trained by maximizing the Evidence Lower Bound (ELBO), $\gL_{\rm ELBO}(\textbf{a}^0, s; \theta)$, which is a lower bound on the log-likelihood $\log \pi_\theta(\textbf{a}^0|s)$. This objective trains the network to reconstruct the clean action $\textbf{a}^0$ from its noised versions $\textbf{a}^n$. For a macro-action $\textbf{a}^0$ with $K$ actions, the objective to maximize is a sum of weighted negative cross-entropy terms over all diffusion steps $n$ and actions $k$:
\begin{equation}
\begin{aligned}
    &\gL_{\rm ELBO}(\textbf{a}^0, s; \theta) = \\
    &\sum_{n=1}^{N} \bar \alpha_n \expect{\textbf{a}^n \sim q(\cdot|\textbf{a}^0)}{ \sum_{k=0}^{K-1} \delta_{a_k^n, m} \cdot \log \mu_\theta(\textbf{a}^n, n, s)_{a_k^0} }
\end{aligned}
\label{eq:elbo_loss}
\end{equation}
where $\bar \alpha_n$ is a weighting term derived from the noise schedule, $\delta_{a_k^n, m}$ is an indicator function that is 1 if the $k$-th action is masked (and 0 otherwise), and $\log \mu_\theta(\cdot)_{a_k^0}$ is the model's predicted log-probability for the original clean action $a_k^0$. The full derivation is detailed in~\Cref{sec.appendix.diffusion}.

%% file: contents/4_algorithm.tex
\section{RL-D$^2$: Reinforcement Learning with Discrete Diffusion}
\label{sec:method}
\input{contents/flowchart_gemini}
We now introduce our framework for training discrete diffusion policies. Our approach follows a policy iteration structure. Let $k$ be the current training iteration. The process alternates between (1) policy evaluation, which estimates the Q-function $q^{\pi_k}$ for the current policy $\pi_k$, and (2) policy improvement. The core of our method lies in the latter improvement step. 

We first define a target distribution, $\pi_{\rm MD}^k$, which is the mirror descent iteration optimal solution from \Cref{eq:opt_pi} calculated using $\pi_{\rm old}\equiv \pi_k$ and $q^{\pi_{\rm old}} \equiv q^{\pi_k}$. This transforms the policy improvement problem into a distributional approximation problem, a common paradigm in deep RL~\citep{chan2022greedification,abdolmaleki2018maximum}. The new policy $\pi_{k+1}$ is then obtained by finding the parameters $\theta$ that minimize a chosen divergence $d$ to this target; namely,
\begin{align}
\pi_{k+1} \in \argmin_{\pi_\theta \in \Pi} \expect*{s \sim \gD}{d\brk{\pi_\theta, \pi_{\rm MD}^k;s}}
\tag{IMPR. STEP}
\label{eq:general_update}
\end{align}
where $\gD$ is a distribution of states, typically from a replay buffer or the current policy's stationary state distribution. {A flowchart summarizing our approach is presented in \cref{fig:rld2_flowchart}.}

\subsection{Reverse and Forward KL}
\label{sec.fkl_and_rkl}

The choice of the divergence $d$ in (\ref{eq:general_update}) is critical and defines the practical update rule. We focus on the Kullback-Leibler (KL) divergence (i.e., $d \equiv d_{KL}$). Specifically, we consider two variants of (\ref{eq:general_update}) using forward KL and reverse KL divergences, which result in two different methods for policy improvement, as we explain below. We refer the reader to~\citep{chan2022greedification} for a thorough review of reverse and forward KL properties in RL.

\paragraph{Forward KL Divergence (FKL).}
The forward KL objective, $d_{KL}(\pi_{\rm MD}^k, \pi_\theta; s)$, seeks a policy $\pi_\theta$ that covers the modes of the target distribution. This "mean-seeking" behavior can be beneficial for exploration, as it encourages the policy to maintain probability mass over all high-value actions~\citep{chan2022greedification}. Minimizing this objective directly is intractable. Instead, we minimize a tractable bound derived by applying the diffusion model's ELBO inequality to the KL definition (see~\Cref{apndx:prop_ce_loss}). This results in the following weighted ELBO loss: 
\begin{align}
&\gL_{\rm FKL}(\theta) = \notag\\
&- \underset{\substack{s \sim \gD \\ \hat{\gA}_s \sim \pi_k}}{\mathbb{E}} \Big[\!\!{\sum_{\textbf{a}^0 \in \hat{\gA}_s} \!\!\!(\underset{\textbf{a} \in \hat{\gA}_s}{\text{softmax}} \brk{A^{\pi_k}(s, \textbf{a}^0) / \lambda} )\cdot \gL_{\rm ELBO}(\textbf{a}^0, s; \theta)} \Big].
\tag{FKL Loss}
\label{eq:fkl_loss}
\end{align}
Here, $\hat{\gA}_s$ is a batch of macro-actions sampled from the "old" policy $\pi_k$ (i.e., a target network, $\pi_{\theta_{\rm old}}$), and {$\gL_{\rm ELBO}$ is the weighted cross-entropy loss in~\eqref{eq:elbo_loss}}. The softmax re-weights the sampled actions to approximate the target distribution $\pi_{\rm MD}^k$. This objective effectively trains the diffusion model as a generative classifier, focusing the model's capacity on reconstructing high-value actions more frequently, and can be easily adapted to off-policy training. 

\paragraph{Reverse KL Divergence (RKL).}
The reverse KL objective, $d_{KL}(\pi_\theta, \pi_{\rm MD}^k; s)$, is equivalent to the original PMD optimization in \Cref{eq:general_pmd} (see~\Cref{apndx:proof_prop_md_eq}). This objective has theoretical policy improvement guarantees~\citep{chan2022greedification} and results in a "mode-seeking" policy that focuses on the highest-value action. This objective can be written as follows:
\begin{align}
&\gL_{\rm RKL}(\theta) =\notag \\
&\expect{s \sim \gD, \textbf{a} \sim \pi_k}{-\eta(s, \textbf{a}; \theta) A^{\pi_k}(s, \textbf{a}) + \lambda d_{KL}(\pi_\theta, \pi_k; s)},
\tag{RKL Loss}
\label{eq:rkl_loss}
\end{align}
where $\eta(s, \textbf{a}; \theta) := \pi_\theta(\textbf{a}|s) / \pi_k(\textbf{a}|s)$ is an importance sampling (IS) ratio. In this case, the usual choice for $\gD$ is the state occupancy measure of $\pi_k$ \citep{schulman2015trust, shani2020adaptive} for on-policy training. 

For diffusion policies, the likelihood $\pi_\theta(\textbf{a}|s)$ is intractable, and thus so is the ratio $\eta$. Following~\citet{ren2024diffusion}, we can construct an augmented MDP where states are $(s, \textbf{a}^n)$ (an environment state and a noisy action at diffusion step $n$) and the advantage for a denoising step is defined by the final clean action's advantage, i.e., $A^{\pi_k}((s, \textbf{a}^n), \textbf{a}^{n-1}) \triangleq A^{\pi_k}(s, \textbf{a}^0)$. This yields a tractable IS ratio based on the single-step reverse process:
$$
\eta((s,\textbf{a}^n), \textbf{a}^{n-1};\theta) = \frac{p_\theta(\textbf{a}^{n-1} | \textbf{a}^n, s)}{p_{k}(\textbf{a}^{n-1} | \textbf{a}^n, s)}.
$$ 
We refer to this ratio as "single-step ratio". Given a tractable estimator for $\eta$, we optimize the objective in \Cref{eq:rkl_loss} using a PPO-style clipping mechanism~\citep{schulman2017proximal}.


\subsection{On-Policy Diffusion Learning}
\label{sec.onpolicy_training}
The standard ELBO objective, used in both \ref{eq:fkl_loss} and \ref{eq:rkl_loss}, trains the model to denoise samples $(\textbf{a}^n, \textbf{a}^0)$ generated from the \emph{fixed forward process} $q(\textbf{a}^n | \textbf{a}^0)$. However, this distribution of noised actions $\textbf{a}^n$ may differ significantly from the actions the policy actually generates during its own generative process. To align the training distribution with the inference distribution, we propose \emph{On-Policy Diffusion Learning}. Instead of starting from a clean action $\textbf{a}^0 \sim \pi_k$ and adding noise, we generate the entire diffusion trajectory $(\textbf{a}^N, \hdots, \textbf{a}^0)$ on-policy by sampling from the \emph{learned reverse process} of the current policy, i.e., $\textbf{a}^N \sim p(\cdot|s)$ and $\textbf{a}^{n-1} \sim p_{\theta_k}(\cdot | \textbf{a}^n, s)$. This yields $(\textbf{a}^n, \textbf{a}^0)$ pairs that are "on-policy" with respect to the policy's own generative dynamics, which we find enhances stability and sample efficiency. {Note that the ``on-policy'' here only refers to the diffusion process, rather than the overall RL framework.}


%% file: contents/flowchart_gemini.tex
\usetikzlibrary{shapes.geometric, arrows, positioning, calc}

\begin{figure*}[t]
\centering
\resizebox{\textwidth}{!}{
\begin{tikzpicture}[
    node distance=0.5cm and 0.5cm
    auto,
    process/.style={rectangle, minimum width=2cm, minimum height=1.2cm, text width=2.2cm, text centered, draw=black, fill=blue!10, rounded corners, font=\small},
    decision/.style={rectangle, minimum width=2cm, minimum height=1.2cm, text width=2cm, text centered, draw=black, fill=green!10, rounded corners, font=\small},
    data/.style={trapezium, trapezium left angle=70, trapezium right angle=110, minimum width=1.5cm, minimum height=1.2cm, text width=1.5cm, text centered, draw=black, fill=orange!10, font=\small},
    collect/.style={ellipse, minimum width=2.5cm, minimum height=1.5cm, text centered, draw=black, fill=yellow!10, font=\small},
    arrow/.style={thick, ->, >=stealth, line width=1pt}
]

\node (buffer) [collect] {Collect Data};

\node (critic) [process, right=0.8cm of buffer] {Policy Evaluation};

\node (qval) [data, right=0.8cm of critic] {$Q$-function\\$q^{\pi_k}$};

\node (pmd) [decision, right=0.8cm of qval] {PMD Target\\$\pi^k_{MD}$};

\node (loss) [process, right=1.2cm of pmd, text width=3.5cm] {Minimize Loss\\$\min_{\pi_\theta} \mathbb{E}[d(\pi_\theta, \pi^k_{MD};s)]$};

\node (update) [data, right=0.8cm of loss] {New Policy\\$\pi_{k+1}$};


\node (diff) [process, above=0.5cm of loss] {Discrete Diffusion Model};


\draw [arrow] (buffer) -- node[midway, above, font=\footnotesize] {$\gD$} (critic);

\draw [arrow] (critic) -- (qval);

\draw [arrow] (qval) -- (pmd);

\draw [arrow] (pmd) -- node[midway, above, font=\footnotesize] {Target} (loss);

\draw [arrow] (diff) -- node[midway, right, font=\footnotesize] {Parameterize $\pi_\theta$} (loss);

\draw [arrow] (loss) -- (update);

\draw [arrow] (update.south) |- ++(0,-1.0) -| (buffer.south) node[pos=0.25, above, font=\footnotesize] {Next Iteration ($k \leftarrow k+1$)};

\node [text width=2.5cm, align=center, font=\footnotesize, above=0.2cm of qval, anchor=south east, xshift=3.5cm] {Construct via Eq. (3)};

\end{tikzpicture}
}
\caption{{\textbf{Overview of the RL-D$^2$ Framework.} Our framework adopts a policy iteration structure. Following policy evaluation, which estimates the current Q-function, the policy improvement step is implemented as a distributional matching problem. Here, the discrete diffusion policy is trained to minimize the KL divergence (FKL or RKL) relative to an optimal target distribution ($\pi_{MD}^k$) derived via Policy Mirror Descent in \eqref{eq:opt_pi}.}}
\label{fig:rld2_flowchart}
\end{figure*}

%% file: contents/5_experiments.tex
\section{Experiments}
\label{sec:experiments}

We conduct a comprehensive set of experiments to evaluate our proposed framework for training discrete diffusion policies. Our evaluation spans three distinct and challenging domains to demonstrate the method's effectiveness, scalability, and versatility: 
\begin{enumerate}
    \item Reward-based finetuning for DNA sequence generation.
    \item Online reinforcement learning with long-horizon decision-making in complex single-agent Atari environments.
    \item Multi-agent cooperative learning with combinatorial joint action spaces.
\end{enumerate}
\subsection{DNA Sequence Generation: Single-Step Policy Optimization}
We first validate our approach on a reward-guided generation task, which serves as a single-step RL problem (i.e., combinatorial multi-armed bandit). The goal is to finetune a pretrained discrete diffusion model to generate DNA sequences that maximize a specific reward signal, verifying the effectiveness of our policy optimization algorithm.

We use a large public enhancer dataset of approximately 700,000 DNA sequences with length 200~\citep{gosai2023machine}. We define a reward function, detailed in Appendix~\ref{sec:dna_setup}, to predict gene expression activity by leveraging the pretrained reward model provided by~\cite{wang2024fine}. Our primary metrics are the reward achieved and the approximate log-likelihood of the generated sequences, which measures their naturalness. We compare against controlled generation methods such as conditional guidance (CG)~\citep{nisonoff2024unlocking}, SMC and TDS~\citep{wu2023practical} and classifier-free guidance (CFG)~\cite{ho2022classifier}, as well as a strong RL-based baseline, DRAKES~\citep{wang2024fine}, that optimizes policies by backpropagating reward through the reverse process using the Gumbel-Softmax trick. For this task, we optimize our policy using the forward KL (FKL) objective~\Cref{eq:fkl_loss}.

\begin{figure}
    \centering
    \includegraphics[width=\linewidth]{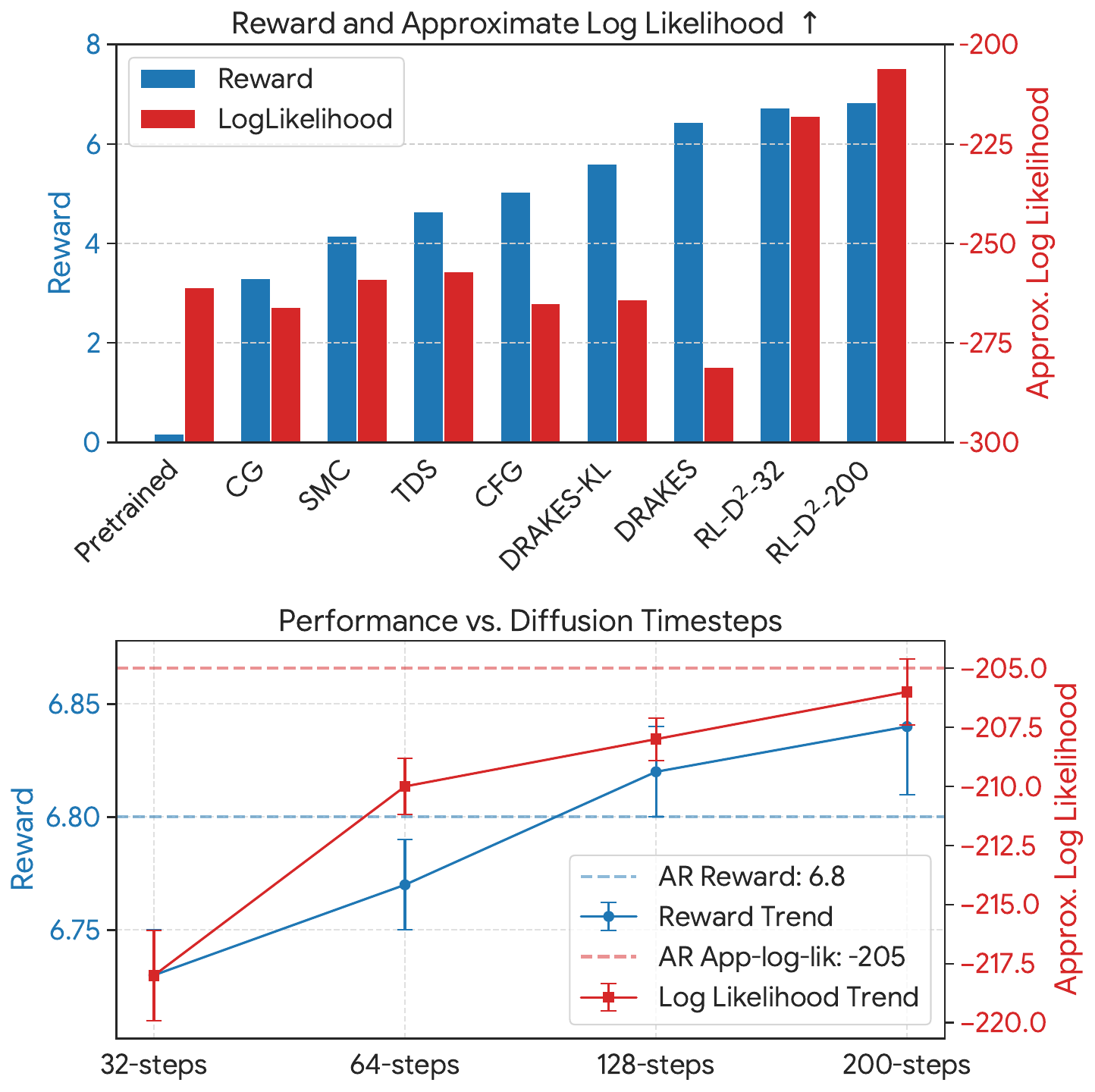}
    \caption{Reward and Approximate Likelihood of DNA generation. \textbf{Left:} The proposed RL-D$^2$ gets best performance on reward and log likelihood, even with fewer diffusion time steps. \textbf{Right:} The mean and 95\% confidence intervals of reward and approximate log likelihood with various diffusion timesteps.}
    \label{fig:dna_generation}
\end{figure}

As shown in~\Cref{fig:dna_generation} (left), our method achieves a new state-of-the-art, attaining the highest reward scores while simultaneously generating the most probable sequences (highest log-likelihood). This demonstrates that our FKL-based update effectively optimizes for the target reward without sacrificing generative quality. Moreover, \Cref{fig:dna_generation} (right) shows that we can achieve consistently strong performance with diffusion timesteps much smaller than the sequence length of 200, highlighting the inference-time efficiency compared to autoregressive~(AR) generation. Finally, our approach is significantly more computationally efficient than DRAKES. Because we do not need to backpropagate through the whole reverse process, we reduce GPU memory consumption from $66.4$~GB to $10.6$~GB, and computation time from $268$ minutes to $97$ minutes compared to DRAKES, making high-performance reward optimization more accessible.

\subsection{Reinforcement Learning with Macro Actions}
\label{sec:main_exp}
Next, we evaluate our RL-D$^2$ in the challenging MinAtar~\citep{young2019minatar} and Atari~\citep{bellemare2013arcade} benchmarks, where the agent learns to make decisions over long horizons by generating macro-actions, i.e., sequences of primitive actions. Our experiments are designed to assess the performance and scalability of using diffusion policies for complex planning tasks. For these tasks, we optimize our policy using the forward KL (FKL) objective in~\Cref{eq:fkl_loss}.

We evaluate RL-D$^2$ on the MinAtar benchmark~\citep{young2019minatar}, a suite of simplified Atari games that provide a controlled setting without partial observability. We employ a \textbf{macro-action length of $4$}. We compare against DQN~\citep{mnih2015human}, IMPALA~\citep{espeholt2018impala}, and their macro-action-enabled variants (\emph{DQN-Macro}, \emph{IMPALA-Macro}). In \emph{DQN-Macro}, the $Q$-network's output dimension is modified to $|\mathcal{A}|^4$ to select one of all possible length-4 macro-actions. For \emph{IMPALA-Macro}, the policy network's output is changed to $4\times |\mathcal{A}|$, allowing it to sample the four actions of the macro-action independently at once. 
The detailed implementations are in~\Cref{sec.appendix.macro_action_baselines}.


\begin{table*}
\centering
\caption{MinAtar performance. Mean and 95\% bootstrap confidence intervals of scores over the last 100 evaluation episodes with 20 random seeds during training on MinAtar.}
\label{tab:mean-scores}
\scshape
\resizebox{\linewidth}{!}{
\begin{tabular}{@{}l|lllll@{}}
\toprule
 & Asterix & Breakout & Freeway & Seaquest & Space Invaders \\
\midrule
DQN
& \textbf{284.38 {\scriptsize [112.88, 455.88]}}
& 180.36 {\scriptsize [64.23, 296.49]}
& \textbf{60.44 {\scriptsize [58.46, 62.42]}}
& 109.20 {\scriptsize [81.23, 137.17]}
& 1.16k {\scriptsize [0.51k, 1.81k]} \\

DQN-Macro
& 43.05 {\scriptsize [32.68, 53.42]}
& 315.85 {\scriptsize [111.90, 519.80]}
& 58.42 {\scriptsize [53.51, 63.33]}
& \textbf{166.65 {\scriptsize [109.25, 224.05]}}
& 769.86 {\scriptsize [258.27, 1281.45]} \\

IMPALA
& 24.78 {\scriptsize [17.38, 32.18]}
& 1.02 {\scriptsize [1.02, 1.02]}
& 41.01 {\scriptsize [26.90, 55.12]}
& 44.55 {\scriptsize [30.95, 58.15]}
& 47.71 {\scriptsize [47.71, 47.71]} \\

IMPALA-Macro
& 21.29 {\scriptsize [12.68, 29.90]}
& 7.21 {\scriptsize [6.78, 7.64]}
& 54.09 {\scriptsize [44.37, 63.81]}
& 55.42 {\scriptsize [32.49, 78.35]}
& 33.18 {\scriptsize [33.18, 33.18]} \\

\rowcolor{shadecolor}
RL-D$^2$ (ours)
& 52.38 {\scriptsize [44.59, 60.17]}
& \textbf{20.58k {\scriptsize [13.47k, 27.46k]}}
& \textbf{59.98 {\scriptsize [57.70, 62.26]}}
& \textbf{164.20 {\scriptsize [109.21, 219.19]}}
& \textbf{184.3k {\scriptsize [62.56k, 305.9k]}} \\
\bottomrule
\end{tabular}
}
\end{table*}

As shown in Table~\ref{tab:mean-scores}, RL-D$^2$ achieves substantially stronger performance in 2 out of 5 tasks in MinAtar. The substantial score improvements in \textsc{Breakout} and \textsc{Space Invaders} mean that the agents can stay alive for a very long time, highlighting the policy's ability to discover and represent complex, long-term strategies. While standard baselines adapted for macro-actions show modest gains, they are far outstripped by our approach, underscoring the necessity of an expressive generative model to effectively navigate large combinatorial action spaces. 

\begin{figure}[!t]
    \centering    \includegraphics[width=\linewidth]{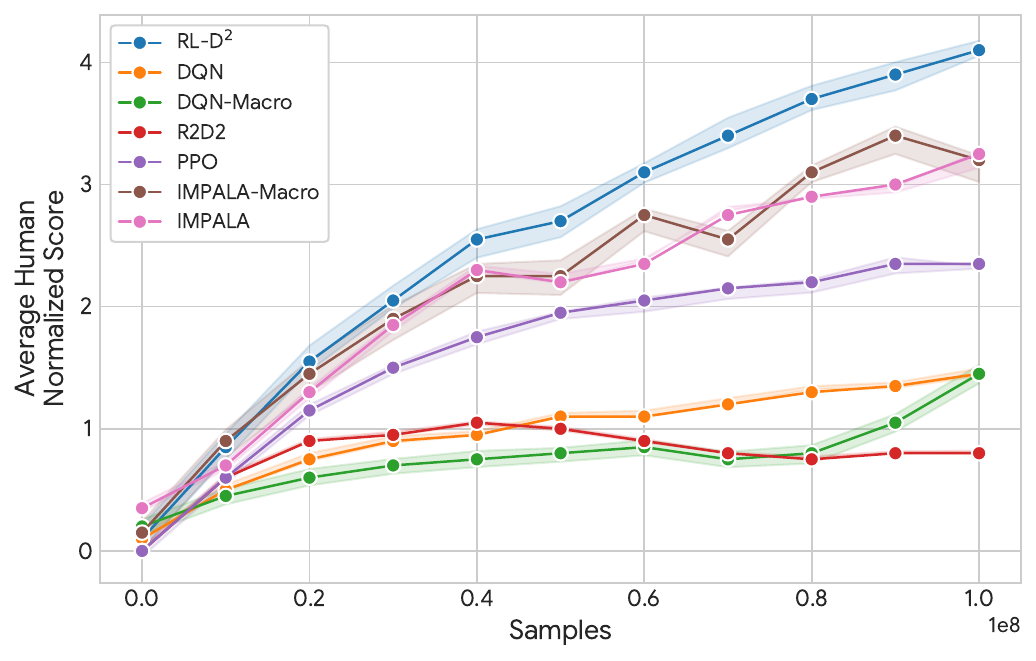}
    \caption{Atari performance.
    Performance improvement over the best baseline, evaluated by the percentage of human normalized scores.
    }
    \label{fig:Atari_performance}
\end{figure}

We confirm our findings on the full Atari benchmark~\cite{bellemare2013arcade} with additional strong baselines including R2D2~\citep{kapturowski2018recurrent} and PPO~\citep{schulman2017proximal}. As shown in Figure~\ref{fig:Atari_performance}, our method achieves the highest human-normalized score on average and outperforms strong baselines using both macro and single actions in 36 of 56 environments (the full results can be found in \Cref{sec.full_results_atari}), showcasing the strong performance of the proposed discrete diffusion. We begin by comparing FKL against RKL to examine which performs better, followed by testing the scalability and horizon-complexity trade-off of our diffusion policy. Lastly, we demonstrate our diffusion policies' performance with different sampling techniques.

\begin{figure}[t]
    \centering
    \includegraphics[width=\linewidth]{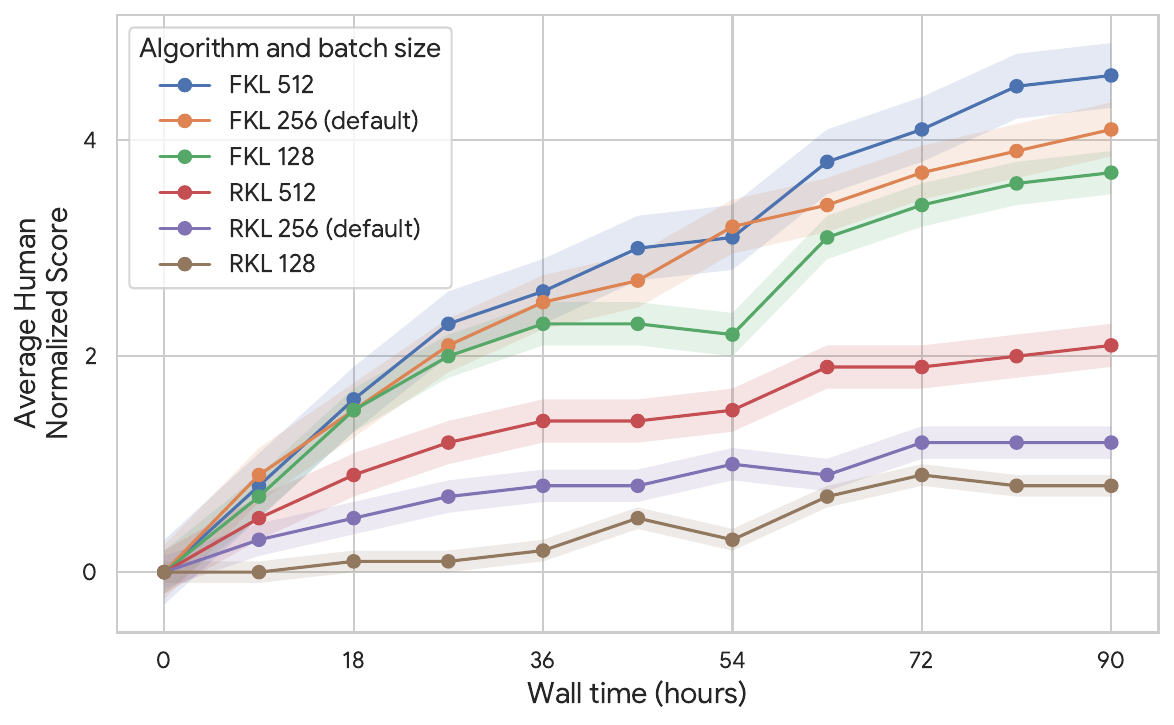}
    \caption{Comparing FKL and RKL, on-policy, in Atari games with different learner batch sizes. The curves show the average human-normalized score over 10 Atari games with 3 random seeds for each game.}
    \label{fig:fkl_rkl_atari}
\end{figure}

\textbf{Comparing FKL vs. RKL with the Same Compute Budget.}
The actual bottleneck in Atari games is computation due to the use of a deep residual network to handle image inputs. Therefore, to compare the performance and stability of FKL and RKL in Atari games, we align the training wall time while adapting the batch size. The results are shown in \Cref{fig:fkl_rkl_atari}. FKL generally outperforms RKL. This observation is re-verified later in the Google Research Football environment (\Cref{sub:marl}). Moreover, FKL is less sensitive to batch size, showing only marginal gains when batch size increases. RKL shows a large performance improvement when increasing batch size from 256 to 512\footnote{One sample in the batch is a 32-step trajectory rather than a transition pair, which is a common practice in training pipelines used for Atari games. Therefore, the actual number of transition pairs is 32 times more.}, showing that RKL favors larger batches because of the high-variance nature of the augmented MDP method.

\begin{figure*}[t!]
    \centering
    \begin{minipage}[b]{0.22\textwidth}
        \centering
        \includegraphics[width=\linewidth]{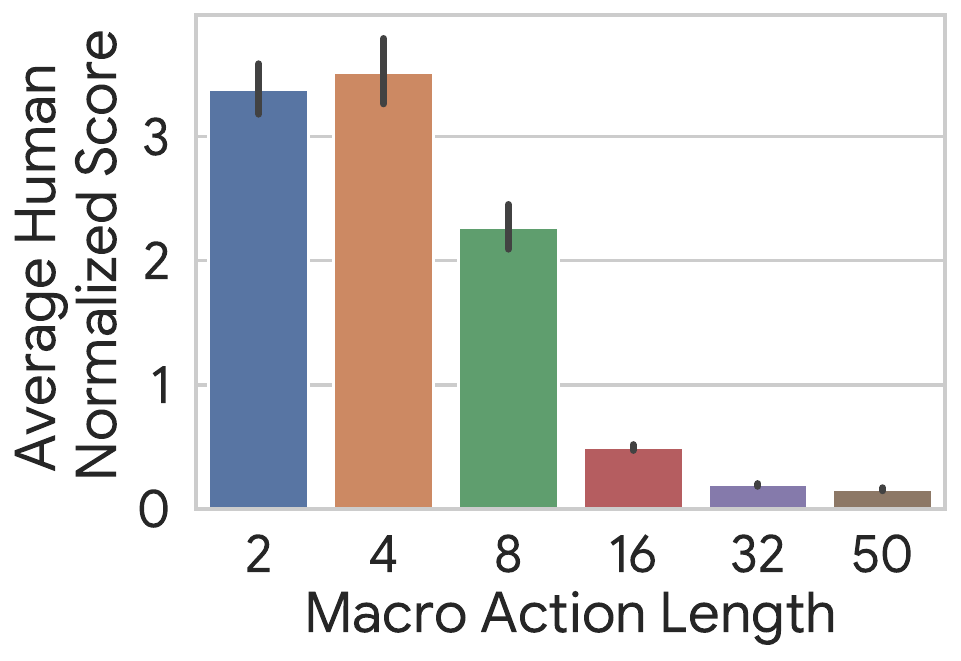}
    \end{minipage}\hfill
    \begin{minipage}[b]{0.22\textwidth}
        \centering
        \includegraphics[width=\linewidth]{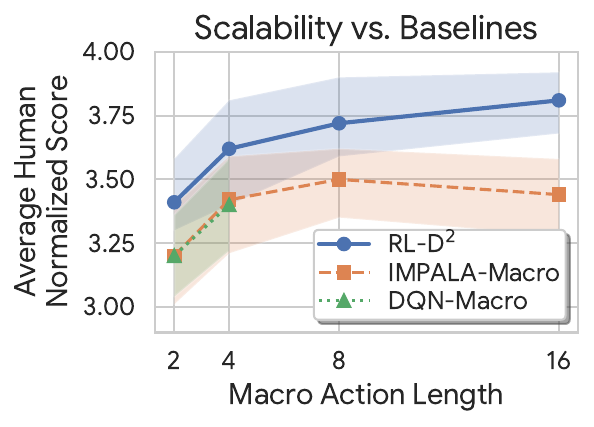}
    \end{minipage}\hfill
    \begin{minipage}[b]{0.54\textwidth}
        \centering
\includegraphics[trim={0cm 0cm 0cm 0.2cm}, clip, width=\linewidth]{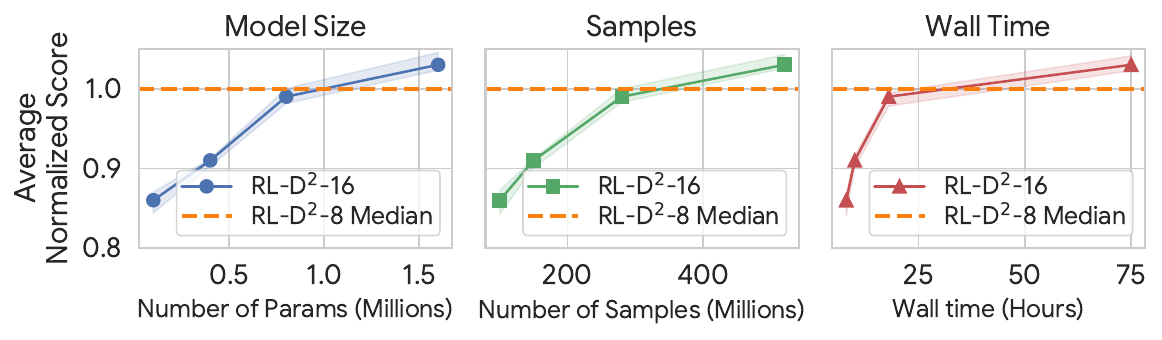}
    \end{minipage}

    \vspace{10pt} 

    \caption{\textbf{Atari Average Human Normalized Score.} \textbf{Left:} Mean and 95\% confidence intervals of averaged episode return over all 56 tasks to show the trade-off between planning horizon and model complexity with fixed network size and data. \textbf{Middle:} The proposed method scales more effectively with increasing network size and data compared to baselines. \emph{DQN-Macro} fails to learn in a reasonable amount of time as the action space grows too large with macro actions more than 4. \textbf{Right:} Mean episode return of RL-D$^2$ with 16 macro actions compared to the 8 macro actions as a function of model parameters, data samples, and training time, averaged over 4 tasks and 3 seed each.}
    \label{fig:trade-off-combined}
\end{figure*}

\textbf{Scalability and horizon-complexity trade-off.} We investigate scalability on more challenging MDPs by varying the macro-action length. The difficulty of solving the MDPs increases because the size of the action space grows exponentially with respect to the macro-action length, while the horizon only shrinks linearly. \Cref{fig:trade-off-combined} shows that with a fixed computational budget, performance peaks at a macro-action length of 4. However, the key advantage of our method is its scalability. As shown in \Cref{fig:trade-off-combined}, when we scale up model capacity and data proportionally to the action space complexity (with setup details listed in~\Cref{sec.appendix.scalable}), our diffusion policy's performance continues to improve; it performs at least as well as IMPALA-Macro up to macro-action length 8, and outperforms IMPALA-Macro at macro-action length 16. \emph{DQN-Macro} cannot fit within reasonable learner resources for macro-action lengths 8 and 16 as the size of the action space increases exponentially, while the baseline \emph{IMPALA-Macro} fails to achieve better performance when the macro-action length increases from 8 to 16. Detailed results for RL-D$^2$ with increasing computational resources from macro-action length 8 to 16 are shown in~\Cref{fig:trade-off-combined}.

\textbf{Efficient and flexible sampling techniques.} We evaluate the inference-time efficiency and flexibility of discrete diffusion policies. To make the difference clear, we extend to a long macro-action setup with length $32$. We leverage two techniques to further improve the sampling quality of discrete diffusion models for these extra-long macro actions: \textbf{(1)} Top-p sampling or nucleus sampling~\citep{holtzman2019curious}, which selects actions from the smallest set of actions whose cumulative probabilities exceed a certain threshold; \textbf{(2)} a remasking diffusion process~\citep{wang2025remasking}, which allows the actions to be masked and unmasked again during the reverse process. The implementation details can be found in \Cref{sec.sampling}.  

\begin{figure}[t!]
    \centering
    \vspace{-10pt}
    \includegraphics[width=0.95\linewidth]{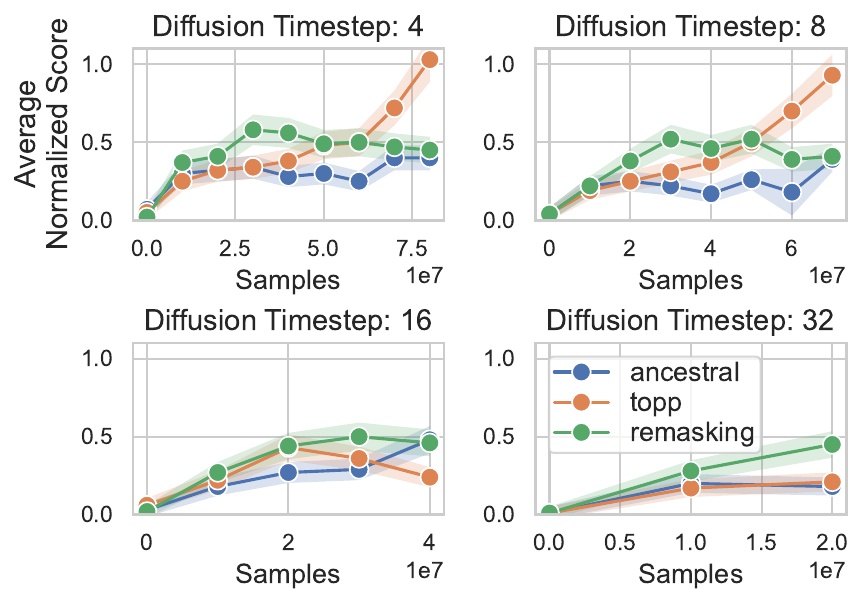}
    \caption{Mean and 95\% confidence intervals of scores averaged over 4 tasks and 3 seed each normalized by best scores achieved with macro action length 32 as a function of diffusion timesteps and sampling techniques. Top-p sampling excels with very few steps while remasking improve the performance with more diffusion steps.}
    \label{fig:sampling}
\end{figure}

As seen in Figure~\ref{fig:sampling}, top-p sampling enhances the performance of RL-D$^2$ with fewer diffusion steps, such as 4 and 8, making inference time more efficient without losing performance. Remasking sampling performs best when the number of timesteps is close to the sequence length. This highlights the flexibility of diffusion models.

\textbf{More Experimental Results and Ablation Studies.}

In the appendix, we explore other techniques and perform ablation studies. This includes \textbf{(1)} automatically tuning the temperature parameter $\lambda$ in~\Cref{eq:opt_pi} by enforcing a hard KL constraint; \textbf{(2)} leveraging discrete diffusion as a planner instead of committing to all macro actions generated, and \textbf{(3)} an ablation study of on-policy diffusion training discussed in~\Cref{sec.onpolicy_training}.
Please refer to~\Cref{sec.appendix.additional_exp} for the results and discussions.

\subsection{Cooperative Multi-Agent Reinforcement Learning}
\label{sub:marl}
Finally, we evaluate our framework in cooperative multi-agent reinforcement learning (MARL), where the combinatorial action space is the joint action of all agents. Efficiently searching this space is a primary challenge in MARL. By modeling the joint action distribution, our diffusion policy can capture complex inter-agent dependencies without relying on restrictive factorization assumptions.

We test our policy on the challenging Google Research Football benchmark~\citep{kurach2020google}, using scenarios ranging from small-scale tasks (\textit{3 vs 1}) to full-team games (\textit{11 vs 11}). Our diffusion model generates the joint action for all controlled agents simultaneously. {We adopt the feature extractor and scenario settings suggested by \cite{song2023boosting}; for further implementation details, please refer to \cref{apndx:grf}. 

We evaluate our diffusion policy, examining both RKL and FKL objective variants against an autoregressive sampler baseline. Additionally, we compare our method against two centralized MARL policy baselines: Multi-Agent PPO (MAPPO) \cite{yu2022surprising} and the \textbf{current state-of-the-art}, Multi-Agent Transformer (MAT) \cite{wen2022multi}. Note that MAT operates autoregressively, mitigating causal bias by training over random permutations of the agents' order. Finally, to assess sample efficiency, we conduct an ablation on the full game scenario (11 vs. 11) using a significantly smaller budget of 100M environment steps (compared to the standard 1G) - marked as \textit{11 vs 11 sm}.}

\begin{table*}[t]
\centering
\caption{Final performance on Google Research Football academy scenarios, measured by maximum mean win rate scaled by 100 with bootstrap confidence intervals over 20 random seeds. Methods with the highest mean and those that have overlapping confidence intervals with the highest mean are \textbf{bolded}.} 
\resizebox{1.\linewidth}{!}{
\renewcommand{\arraystretch}{1.7}
\begin{tabular}{c|cccccccc|c}
\toprule
Method
& Pass \& Shoot
& Run Pass \& Shoot
& 3 vs 1
& CT-Easy
& CT-Hard
& Corner
& 5 vs 5
& 11 vs 11
& 11 vs 11 SM \\
\midrule
Autoregressive
& 79.4 {\small [44.2, 94.5]}
& 68.9 {\small [36.4, 89.3]}
& 74.1 {\small [43.4, 91.7]}
& 70.8 {\small [43.0, 89.5]}
& 35.2 {\small [0.5, 66.5]}
& 0.0 {\small [0.0, 0.0]}
& 72.5 {\small [46.0, 90.1]}
& 33.1 {\small [19.8, 46.4]}
& 0.0 {\small [0.0, 0.0]} \\

\rowcolor{shadecolor}
RL-$\mathrm{D}^2$ RKL (ours)
& \textbf{100.0} {\small \textbf{[100.0, 100.0]}}
& \textbf{97.9} {\small \textbf{[96.3, 99.1]}}
& \textbf{99.2} {\small \textbf{[98.5, 99.7]}}
& \textbf{98.5} {\small \textbf{[97.7, 99.2]}}
& \textbf{98.2} {\small \textbf{[96.3, 99.4]}}
& \textbf{97.1} {\small \textbf{[96.3, 98.9]}}
& \textbf{99.1} {\small \textbf{[98.7, 99.5]}}
& \textbf{99.4} {\small \textbf{[99.0, 99.7]}}
& 0.0 {\small [0.0, 0.0]} \\

\rowcolor{shadecolor}
RL-$\mathrm{D}^2$ FKL (ours)
& \textbf{98.4} {\small \textbf{[97.3, 99.3]}}
& \textbf{99.2} {\small \textbf{[98.7, 99.6]}}
& 96.8 {\small [94.8, 98.5]}
& \textbf{97.4} {\small \textbf{[96.2, 98.6]}}
& \textbf{96.5} {\small \textbf{[93.7, 98.6]}}
& 94.1 {\small [91.7, 96.1]}
& 97.9 {\small [96.8, 98.5]}
& 98.1 {\small [97.1, 99.0]}
& \textbf{68.5} {\small \textbf{[62.0, 75.0]}} \\

MAT
& \textbf{96.8} {\small \textbf{[94.6, 98.6]}}
& \textbf{97.6} {\small \textbf{[96.3, 98.8]}}
& 91.8 {\small [90.6, 93.0]}
& 88.5 {\small [86.4, 90.6]}
& 87.1 {\small [83.5, 90.7]}
& 94.6 {\small [92.1, 95.9]}
& 92.9 {\small [91.9, 93.9]}
& 91.3 {\small [88.9, 93.7]}
& 10.2 {\small [7.4, 13.0]} \\

MAPPO
& 98.8 {\small [98.4, 99.2]}
& 74.5 {\small [71.6, 77.4]}
& 92.6 {\small [91.0, 94.2]}
& 71.4 {\small [68.3, 74.5]}
& 64.2 {\small [62.1, 66.3]}
& 54.7 {\small [50.2, 59.2]}
& 94.8 {\small [93.1, 96.5]}
& 53.9 {\small [50.9, 56.9]}
& 4.8 {\small [4.5, 5.1]} \\
\bottomrule
\end{tabular}
}
\label{table:gfootball}
\end{table*}

{As presented in Table~\ref{table:gfootball}, our discrete diffusion policy variants achieve the highest mean win rates across all scenarios. This advantage is particularly pronounced in the most challenging tasks requiring intricate team coordination, such as \textit{5 vs 5}, \textit{corner}, \textit{ct-hard}, and the highly complex \textit{11 vs 11} full game, where our method shows a clear advantage over state-of-the-art methods.
Regarding the baselines, we note that the naive autoregressive sampler lags significantly behind. 

This suggests that to match the performance of diffusion models, autoregressive methods require additional alignment techniques, such as the random permutations used in MAT, to mitigate causal bias. Furthermore, while the RKL objective yields superior results overall, FKL significantly outperforms RKL (and other baselines) in the small-budget regime \textit{11 vs 11 sm}, approaching a 70\% win rate. RKL is more exploratory, allowing it to achieve optimal behavior given sufficient samples, whereas FKL exploits faster at the cost of less exploration. Moreover, we observe that FKL is likely to collapse if the temperature tuning is not handled well (results shown in \Cref{sec.appendix.temp.gft}). Overall, these results highlight the potential of discrete diffusion models to effectively generate highly coordinated joint actions in challenging multi-agent tasks.}

{\subsection{Empirical Selection of FKL and RKL}

We summarize our empirical observations about FKL and RKL to provide guidance to practitioners from the viewpoint of balancing the trade-off between computational efficiency and asymptotic performance.

\textbf{Data efficiency}: FKL demonstrates faster learning in the initial stages with fewer samples, as evidenced by the \textit{11 vs 11 sm} scenario in~\Cref{table:gfootball} and the Atari benchmarks in~\Cref{fig:fkl_rkl_atari}. We attribute this to the temperature values employed. As shown in \Cref{fig:appendix_temp}, FKL performs well only with a large KL constraint and small temperatures; this configuration promotes aggressive exploitation by rapidly shifting the policy toward being greedy and deterministic. In contrast, our RKL implementation utilizes importance sampling ratio clipping. This mechanism results in more gradual shifts in the policy distribution, thereby leading to a slower learning pace.

\textbf{Asymptotic performance and stability.} \Cref{table:gfootball} demonstrates that RKL achieves comparable or superior performance to FKL when trained with large sample sizes. Moreover, the temperature schedule ablation for Google Football in \Cref{fig:appendix_temp_gft} indicates that FKL is prone to collapse if the temperature is not appropriately selected. This slightly inferior performance of FKL may stem from two factors: (1) FKL optimizes the ELBO, a lower bound of policy loss, whereas RKL optimizes the unbiased policy mirror descent loss; and (2) the potential for collapse is caused by the policy becoming too deterministic, resulting in a lack of diversity in the replay buffer.

In summary, our empirical results show that RL-D$^2$-FKL learns faster through heavy exploitation, but is less stable and requires careful temperature tuning. It is suitable for tasks with data and computational bottlenecks such as Atari games. RL-D$^2$-RKL learns slightly slower but is more stable and has better asymptotic performance. It is suitable for tasks with cheap sampling costs, such as Google Football.
}

%% file: contents/6_conclusion.tex
\section{Summary}
This paper introduces RL-D$^2$, a novel framework for reinforcement learning with discrete diffusion policies, aimed at solving decision making problems with large, combinatorial action spaces. In this framework, we propose to train diffusion models by fitting their output distribution to the analytic solution of the Policy Mirror Descent policy optimization algorithm. This is done by projecting the outputs of the model with either the forward or reverse KL divergences. Extensive experiments demonstrate that the proposed method achieves state-of-the-art performance across three challenging domains: reward-guided sequence generation, long-horizon planning with macro-actions, and cooperative multi-agent RL. The results suggest that the RL-D$^2$ framework provides a scalable and high-performing solution to a long-standing challenge in RL, effectively handling complex, combinatorial action spaces where traditional methods often fail.

\section*{Impact Statement}
This paper presents work whose goal is to advance the field of machine learning. There are many potential societal consequences of our work, none of which we feel must be specifically highlighted here.




%% file: contents/A_appendix.tex
    \section{Masked Discrete Diffusion Process}
\label{sec.appendix.diffusion}



Like continuous diffusion, discrete diffusion is composed of a fixed forward process and a learned reverse process. The forward process degrades a data sample $x^0\sim q\left(x^0\right)$ into a sequence of progressively noisier latent variables $x^1, x^2, \ldots, x^N$ via a Markov chain, 
$$
q\left(x^{1: N} \mid x^0\right)=\prod_{n=1}^N q\left(x^n \mid x^{n-1}\right), \text{where}\quad q\left(x^n \mid x^{n-1}\right)=\operatorname{Cat}\left(x^n ; p=\mathbf{Q}_n^\top x^{n-1}\right)
$$
where $\mathbf{Q}_n$ is the transition matrix with $
\left[\mathbf{Q}_n\right]_{i j}=q\left(x^n=j \mid x^{n-1}=i\right)
$. 
Specifically, we focus on a family of discrete diffusion processes called masked diffusion models~\citep{austin2021structured,campbell2022continuous,shi2024simplified}, where an additional \texttt{[MASK]} action is added to the action space. Denoting the mask action as a special $\texttt{m}$ action, the transition kernel of the forward process is defined as 
$$
\mathbf{Q}_n = (1-\beta_n)\mathbf{I}+ \beta_n\mathbf{1} \mathbf{e}_m^\top,
$$ 
In other words,
$$
q\left(x^n \mid x^{n-1}\right) = \begin{cases}
    \beta_n & \text{if } x^{n-1}\neq\texttt{m} \text{ and } x^n=\texttt{m}\\
    1-\beta_n  & \text{if } x^{n-1}\neq\texttt{m} \text{ and } x^n=x^{n-1}\\
    1 & \text{if } x^{n-1}=x^n=\texttt{m}\\
    0 &\text{otherwise}
\end{cases}
$$
which means the action is masked out with probability $\beta_n$ and otherwise stays the same. The masking schedule is defined as $\alpha_n := \prod_{i=1}^n (1-\beta_i)$. Once the action is masked, it stays as the masked action $\texttt{m}$.
The learned reverse Markov process $p_\theta (x^{0:N}) = p(x^N) \prod_{n=1}^N p_\theta (x^{n-1}|x^n)$ gradually denoises (unmasks) the latent variables
towards the data distribution. The reversal model estimates the posterior:
$$
q(x^{n-1} | x^n, x^0) = \begin{cases} \text{Cat}\left(x^{n-1}; \bar \alpha_n x^0 + (1 - \bar \alpha_n)\textbf{e}_m \right) & \quad x^n = \textbf{e}_m, \\
\text{Cat}\left(x^{n-1}; x^n \right) & \quad x^n \neq \textbf{e}_m
\end{cases}
$$ 
We parameterize this posterior using $p_\theta(x^{n-1} | x^n) := q(x^{n-1}|x^n,  \mu_\theta(x^n,n))$, where $\bar \alpha_n := \frac{\alpha_{n-1} - \alpha_n}{1 - \alpha_n}$ and
\[
\mu_\theta(x^n, n) = \begin{cases}
\text{softmax}(f_\theta(x^n, n)) & \quad x^n = \texttt{m}, \\
x^n & \quad x^n \neq \texttt{m}.
\end{cases}
\]

Here, $\mu_\theta$ is the clean sample mean-value estimator induced by a trained model $f_\theta$ optimized by maximizing the Evidence Lower Bound (ELBO)
\begin{equation}
    \gL_{\rm ELBO}(x^0; \theta) = \sum_{n=1}^N \bar \alpha_n \expect*{x^n \sim q_{n|0}}{\delta_{x^n, \texttt{m}} \cdot (x^0)^\top \log\mu_\theta(x^n, n)},
\label{eq:elbo}
\end{equation}
where $\delta_{x, y}$ is an indicator function and $q_{n|0} := q(x^n | x^0)$. The ELBO acts as a lower bound for the expected log-likelihood
\begin{equation}
    \log p_\theta(x^0) \geq \gL_{ELBO}(x^0; \theta).
\end{equation}
Throughout this work, we abuse the notation such that $x^n$ can be either an integer or its corresponding one-hot vector, whenever it is clear from the context.

\section{Proofs and Derivations}
\label{sec.appdendix.derivations}
\subsection{Proof to the Forward KL Loss~\Cref{eq:fkl_loss}}
\label{apndx:prop_ce_loss}
\begin{proof}
Denoting the forward KL:
\begin{align*}
  d_{KL}(\pi_{MD}, \pi_\theta;s) &= \sum_{\textbf{a} \in \gA^K} \pi_{MD}(\textbf{a}|s) \log \frac{\pi_{MD}(\textbf{a} | s)}{\pi_\theta(\textbf{a} | s)} \\
  &= \sum_{\textbf{a} \in \gA^K} \pi_{MD}(\textbf{a}|s) \log \pi_{MD}(\textbf{a} | s) - \sum_{\textbf{a} \in \gA^K} \pi_{MD}(\textbf{a}|s)  \log \pi_\theta(\textbf{a} | s) \\
  &= -\gH(\pi_{MD}(\cdot|s)) - \sum_{\textbf{a} \in \gA^K} \pi_{\rm old}(\textbf{a}|s) \frac{\exp \brk{q^{\pi_{\rm old}}(\textbf{a},s) / \lambda}}{Z(s)} \log \pi_\theta(\textbf{a} | s) \\
  &\leq - \sum_{\textbf{a} \in \gA^K} \pi_{\rm old}(\textbf{a}|s) \frac{\exp \brk{q^{\pi_{\rm old}}(\textbf{a},s) / \lambda}}{Z(s)} \gL_{ELBO}(\textbf{a},s;\theta) \\
  &= \expect*{\textbf{a} \sim \pi_{\rm old}}{- \frac{\exp \brk{q^{\pi_{\rm old}}(\textbf{a},s) / \lambda}}{Z(s)} \gL_{ELBO}(\textbf{a},s;\theta)},
\end{align*}

where the first equality is the KL divergence definition, the third equality comes from the definition of entropy $\gH$ and the definition of the MD policy in \cref{eq:opt_pi}, and the inequality comes from the ELBO inequality w.r.t.\ $\log\pi_\theta(\textbf{a} |s)$ of the discrete diffusion policy and from the fact that entropy is non-negative.  
\end{proof}

That is, we can bound the forward KL metric using the self-imitation objective. Given the support of $\pi_{\rm old}$: $\gA(\pi_{\rm old};s) = \brk[c]{\textbf{a}^0 | \pi_{\rm old}(\textbf{a}^0 |s) > 0, \textbf{a}^0 \in \gA^k}$, the self-imitation loss is a weighted average of the discrete diffusion loss with softmax weights $w_\lambda$ over this support. The SI loss is a generalized version of the classification policy iteration \cite{lazaric2016analysis}, which is the solution w.r.t.\ the non-regularized MD policy: 

\begin{remark}[Classification Discrete Diffusion Loss] 
    Consider the limit MD policy's temperature $\lambda \rightarrow 0$,  we get that:
    $$
    \lim_{\lambda \rightarrow 0} \expect*{\textbf{a} \sim \pi_{\rm old}}{- \frac{\exp \brk{q^{\pi_{\rm old}}(\textbf{a},s) / \lambda}}{Z(s)} \gL_{ELBO}(\textbf{a},s;\theta)} = - \frac{1}{n^*} \sum_{\textbf{a}^* \in \gA^*(\pi_{\rm old};s)} \gL_{ELBO}(\textbf{a}^*, s; \theta),
    $$
    where $\gA^*(\pi;s) := \argmax_{\textbf{a} \in \gA(\pi_{\rm old};s)} q^\pi(\textbf{a},s) $ and $n^* := |\gA^*(\pi;s)|$. This is a classification policy iteration \citep{lazaric2016analysis} of the discrete diffusion policy.
\end{remark}

Overall, the self-imitation loss encapsulates a weighted behavioral cloning objective w.r.t.\ the MD policy. In practice, computing $w_\lambda$ is non-trivial, as sampling actions from the whole action space may be expensive, especially in the domains considered in this work. Therefore, a practical approach would be to estimate $w_\lambda$ by sampling a subset of actions $\hat \gA_s \sim \pi_{\rm old}(\cdot | s)$, where $\hat \gA_s := \brk[c]{\textbf{a}_i}_{i=1}^M, \textbf{a}_i \sim \pi_{\rm old}(\cdot | s)$ and perform a softmax over their $q$-function values, which effectively estimates the normalization factor over the sampled set:
\begin{equation*}
    Z(s) \approx \frac{1}{M} \sum_{\textbf{a} \in \hat \gA_s} \exp \brk{{q^{\pi_{\rm old}}(s, \textbf{a}})/\lambda} = \frac{\hat Z(s)}{M}.
\end{equation*} 
This gives us the next approximated loss:
\begin{equation*}
    \gL_{FKL}(\theta) = - \expect*{s \sim \gD, \hat{\gA}_s \sim \pi_{\rm old}}{ \sum_{\textbf{a}^0 \in \hat\gA_s}\hat w_\lambda(s, \textbf{a}^0) \gL_{ELBO}(\textbf{a}^0, s; \theta)},\label{eq:loss_ce}
\end{equation*}

where $\hat w_\lambda(s, \textbf{a}) := \frac{\exp \brk{{q^{\pi_{\rm old}}(s, \textbf{a}})/\lambda}}{\hat Z(s)}$ and $\gD$ is the replay buffer.

\subsection{Proof to The Reverse KL Loss~\Cref{eq:rkl_loss}}
\label{apndx:proof_prop_md_eq}
\begin{proof}
Denoting the reverse KL:
\begin{align*}
d_{KL}(\pi_\theta, \pi_{MD} ;s) &= \sum_{\textbf{a} \in \gA^K} \pi_{\theta}(\textbf{a}|s) \log \frac{\pi_{\theta}(\textbf{a} | s)}{\pi_{MD}(\textbf{a} | s)} \\
&= \sum_{\textbf{a} \in \gA^K} \pi_{\theta}(\textbf{a}|s) \log \pi_{\theta}(\textbf{a} | s) - \sum_{\textbf{a} \in \gA^K} \pi_{\theta}(\textbf{a}|s) \log \pi_{MD}(\textbf{a} | s) \\
&= \sum_{\textbf{a} \in \gA^K} \pi_{\theta}(\textbf{a}|s) \log \pi_{\theta}(\textbf{a} | s) - \sum_{\textbf{a} \in \gA^K} \pi_{\theta}(\textbf{a}|s) \log \pi_{\rm old}(\textbf{a} | s) - \lambda^{-1} \sum_{\textbf{a} \in \gA^K} \pi_{\theta}(\textbf{a}|s) q^{\pi_{\rm old}}(\textbf{a},s) + Z(s) \\
& = d_{KL}(\pi_\theta, \pi_{\rm old} ;s) -\lambda^{-1} \expect*{\textbf{a} \sim \pi_\theta}{q^{\pi_{\rm old}}(\textbf{a},s)} + Z(s)
\end{align*}
Since the normalization factor is independent of $\theta$
$$
\argmin_\theta  d_{KL}(\pi_\theta, \pi_{MD} ;s) = \argmax_\theta \expect*{\textbf{a} \sim \pi_\theta}{q^{\pi_{\rm old}}(\textbf{a},s)} - \lambda d_{KL}(\pi_\theta, \pi_{\rm old} ;s)
$$
which is the mirror descent objective regularized with $\pi_{\rm old}$.
\end{proof}





\section{Implementation Details}

\label{sec.appendix.implementations}
\subsection{RL-D$^2$ Implementation Details for Atari}

 We follow similar off-policy distributed RL framework as R2D2~\citep{kapturowski2018recurrent} implemented on ACME~\citep{hoffman2020acme}. In Atari games, we leverage the same recurrent feature extraction in~\citep{kapturowski2018recurrent} by unrolling an LSTM network. We leverage the priority experience replay~\citep{horgan2018distributed}. The hyperparameters are listed in~\Cref{tab:atari_hyperparameters}.
\begin{table}[h]
\centering
\caption{RL-D$^2$ Hyperparameters with FKL}\label{tab:atari_hyperparameters}
\begin{tabular}{@{}llll@{}}
\toprule
Parameters                & Value                        & Parameters                & Value \\ \midrule
Number of samples         & 1e8                          & Sample-to-insert Ratio    & 4.0   \\
Number of parallel actors & 16                           & Mini Batch size                & 64  \\
 Unroll length& 40 & Burn-in length & 8 \\
$Q$-network             & Transformer & Target update rate &  0.005     \\
Policy network            & Transformer & Actor learning rate &  $1\times 10^{-4}$     \\
Tempearture learning rate           & $1\times 10^{-2}$ & Critic leaning rate &  $1\times 10^{-4}$    \\
Transformer hidden dim            & 80 & Transformer layers &  3     \\
Transformer heads            & 1 & Discounted factor & 0.997 \\
Priority exponent & 0.99& Replay buffer size & $5\times 10^6$ \\
\bottomrule
\end{tabular}
\end{table}

\textbf{Model Architectures.} 
The same $3$-layer convolutional network structure as \cite{kapturowski2018recurrent,mnih2015human} is used for all the algorithms, followed by an LSTM with 512 hidden units, which feeds into actor and value networks implemented as transformers.

The learnable parameters inside our transformer include the input embedding, linear projections for conditioning, weights/biases in the multi-head attention and feed-forward networks within each transformer block. Key learnable parameters are also the adaptive normalization layers (lns) that generate dynamic shift, scale, and gate values based on the conditioning. Finally, the output projection is learnable.
Conditioning is introduced via a FiLM-like (Feature-wise Linear Modulation) mechanism. The objective then passes through small linear networks to produce shift, scale, and gate parameters. These dynamically modulate activations after layer normalization in both the attention and feed-forward sub-layers. For example, inputs to sub-layers become \texttt{norm(h) * (scale + 1.0) + shift}, with gate controlling residual connections. This enables the transformer to adapt its internal computations layer-wise based on external conditions.

\subsection{Temperature Tuning By Hard KL Divergence Constraints}
\label{sec.appendix.temp_tuning}
Autotuning the temperature parameters $\lambda$ has been a challenging problem for RL with diffusion policies, as the output log probabilities are unknown. Existing methods leverage Gaussian mixture fitting~\cite{wang2024diffusion}, uniform data insertion~\cite{ding2024diffusion}, and data processing inequalities~\cite{celik2025dime}.

We noticed a simple approach using duality by enforcing a hard constraint on the KL divergence based on~\citet{abdolmaleki2018maximum}. Consider the policy mirror descent with hard constraints,
\begin{equation}
    \begin{aligned}
        \max_{\pi} &\expect*{\textbf{a}\sim\pi }{A^{\pi_{\rm old}}(s, \textbf{a})}\\
        \text{s.t.}~& d_{\rm KL}(\pi, \pi_{\rm old})\leq \epsilon
    \end{aligned}
\end{equation}
To solve it, we construct the Lagrangian

$$
L(\pi, \lambda, \eta) = \expect*{a\sim \pi}{A^{\pi_{\rm old}} (s, \textbf{a})} + \lambda (\epsilon - d_{\rm KL}(\pi, \pi_{\rm old})) + \eta (1-\sum_a\pi(\textbf{a}|s) )
$$

Gradient to the primal objective,
$$
\partial \pi L = A^{\pi_{\rm old}}(s, \textbf{a}) - \lambda (\log \frac{\pi(\textbf{a}|s)}{\pi_{\rm old}(a|s) } + 1) + \eta
$$

Let it equal $0$ we get the primal optimal solution is 
$$
\pi = \pi_{\rm old }(a|s)\exp(\frac{A^{\pi_{\rm old}}(s, \textbf{a})}{\lambda}) \exp(\frac{\eta - \lambda}{\lambda})
$$

As we have the normalization constraint, we have

$$
\exp(-\frac{\eta - \lambda}{\lambda}) = \sum_{\textbf{a}} \pi_{\rm old }(a|s)\exp(\frac{A^{\pi_{\rm old}}(s, \textbf{a})}{\lambda}) := Z
$$
Therefore, we have
$$
\eta = \lambda(1-\log Z)
$$
Substituting this back to the Lagrangian, we have
$$
\begin{aligned}
    g(\lambda) = & \lambda\epsilon + \eta + \sum_{\textbf{a}}\pi(\textbf{a}|s) \left(A^{\pi_{\rm old}}(s, \textbf{a}) - \lambda\log \left(\exp (A^{\pi_{\rm old}}(s, \textbf{a})/\lambda)\right) +\lambda\log Z - \eta\right)\\
    = & \lambda\epsilon + \sum_{\textbf{a}}\pi(\textbf{a}|s) (\lambda \log Z) \\
    = & \lambda\epsilon +  \lambda \log Z\\
    = & \lambda\epsilon +  \lambda \log \sum_{\textbf{a}} \pi_{\rm old }(a|s)\exp(\frac{A^{\pi_{\rm old}}(s, \textbf{a})}{\lambda})\\
    \approx & \lambda \epsilon + \lambda  \log \frac{1}{N}\sum_{i=1}^N\exp(\frac{A^{\pi_{\rm old}}(s, \textbf{a}_i)}{\lambda}) & a_i\sim \pi_{\rm old }(\textbf{a}_i|s) \\
    = & \lambda\epsilon + \lambda\operatorname{logsumexp}(\frac{A^{\pi_{\rm old}}(s, \textbf{a}_i)}{\lambda}) - \lambda \log N
\end{aligned}
$$

Therefore, we can update the temperature parameter by $\min g(\lambda)$, which can be used in discrete and continuous diffusion.

\subsection{Sampling Techniques.}
\label{sec.sampling}
\begin{itemize}
    \item \textbf{Top-p sampling.} For each action that is sampled to be unmasked, we select the smallest set of actions whose cumulative probability computed from $f_\theta$ exceeds $P=0.98$. Then we re-normalize the distribution including only these actions and sample from this re-normalized distribution.
    \item \textbf{Re-masking.} Using the same techniques as~\cite{wang2025remasking}, we do not need to change the ELBO~\Cref{eq:elbo} or the FKL policy loss~\Cref{eq:fkl_loss}. We only need to change the inference-time sampling procedure.
\end{itemize}

\section{Experiments}
\label{sec.appendix.experiments}
\subsection{DNA Generation Setup}
\label{sec:dna_setup}
\textbf{Dataset.} The experiment is based on a large, publicly available dataset of enhancers, which contains activity measurements for approximately 700,000 DNA sequences, each 200 base pairs long, within human cell lines. The dataset contains the expression level, which is also used to train our reward models. A masked discrete diffusion model was pretrained on the complete set of sequences. 

\textbf{Reward models.} Following established conventions in~\citep{wang2024fine}, the dataset was then divided by chromosome to train two distinct reward models, or "oracles". These oracles, built on the Enformer~\citep{avsec2021effective} architecture, were designed to predict the enhancer activity level in the HepG2 cell line; one oracle was used to fine-tune the models, while the other was reserved for evaluation.

\textbf{Evaluation Metrics.} To conduct a thorough assessment of each model's ability to generate effective enhancers, the following metrics were employed:

\begin{enumerate}
    \item Predicted Activity (Reward): This metric measures the enhancer activity level in the HepG2 cell line as predicted by the evaluation reward oracle. It's important to note that the models were fine-tuned using a separate oracle trained on a different chromosomal subset of the data.

    
    
    
    \item Approximated Log-Likelihood (App-Log-Lik): The log-likelihood of the generated sequences was calculated with respect to the pretrained model. This measures how "natural" the sequences are; a low likelihood would indicate that the model over-optimized the reward and generated out-of-distribution sequences.
\end{enumerate}

\subsection{Implementation of the \textbf{Macro} baselines}
\label{sec.appendix.macro_action_baselines}
We used the following adaptations to convert baseline algorithms to the setup with macro actions.
\begin{itemize}
    \item \textbf{DQN-Macro: } For DQN, we directly make the $Q-$network output a vector of size $|\mathcal{A}|^K$, which is the size of the total combinatorial action space.
    \item \textbf{IMPALA-Macro: } Instead of outputting $|\mathcal{A}|$ logits for a single action, the actor networks predict $K\times|\mathcal{A}|$ logits, where each $|\mathcal{A}|$-dimensional vector gives the logits for one action in the macro-action.
\end{itemize}

{

\subsubsection{Hyperparameter Search for the DQN-Macro}
We conduct hyperparameter search for the baseline \textbf{DQN-Macro} algorithm to test whether our macro baselines prefer different hyperparameters from the original algorithm. We search over the following 3 parameters:

\begin{itemize}
    \item MLP layers. The original paper includes a 1-layer MLP head over the deep residual feature extractors. We increase the number of layers to \textbf{2 and 3 } to see whether the increased number of parameters benefits macro actions. 
    \item Prioritized experience replay exponents. Current one used for DQN is 0.6 and we try 0.4.
    \item The value of $\epsilon$ in the $\epsilon$-greedy exploration. Current one used for DQN is 0.01 and we try 0.001 and 0.1.
\end{itemize}

The results are shown in \Cref{fig:dqn_hyperparameters}, which shows that the default DQN parameters, also the one we used in our main results, still perform the best for the DQN with macro actions. 
\begin{figure}[h]
    \centering
    \includegraphics[width=\linewidth]{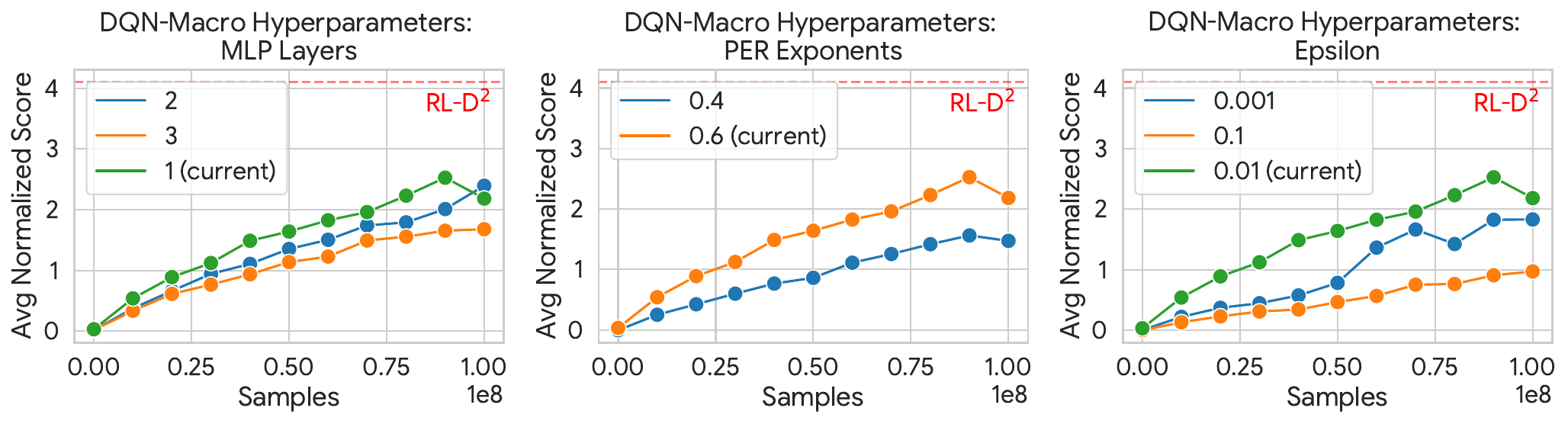}
    \caption{DQN-Macro hyperparameter search. The results shows averaged human normalized score over 10 Atari environments, each with 3 random seeds. Red dash line is the proposed RL-D$^2$ for reference. }
    \label{fig:dqn_hyperparameters}
\end{figure}
}
\subsection{Scalability and Horizon-Complexity Trade-off Setup}
\label{sec.appendix.scalable}
Compared to the hyperparameters in \Cref{tab:atari_hyperparameters}, we scale the computation under the following conditions:
\begin{itemize}
    \item \textbf{Number of environment steps:} $5\times 10^8$ maximum for macro action length 8 and 16.
    \item \textbf{Model sizes:} We change the heads of the transformers: $2$ heads for macro-action length $4$, $4$ heads for macro-action length $8$, and $6$ heads for macro-action length 16.
    \item \textbf{Batch size.} We change the mini-batch size to 128, resulting in a total effective batch size of 4096 for macro action length $8$ and $16$.
\end{itemize}

\subsection{RL-D$^2$ Implementation of Google Research Football}
\label{apndx:grf}
For Google Research Football, we used the same features as in \cite{song2023boosting} and the same scenario settings for training and evaluation. We implemented the RKL version of RL-$\text{D}^2$ with a single-step ratio in a PPO \citep{schulman2017proximal} framework. For the state embeddings, we used an additional transformer similar to the one used for the diffusion process. The transformer outputs a state embedding for each player, which is fed as a condition for the action transformer and also fed to a value-head MLP network that outputs a value for each player. The value is trained as described in \cite{wen2022mat}. {We trained the model and baselines over 1G environment steps for \textit{11 vs 11} and 500M for the rest of the scenarios.}

\begin{table}[h]
\centering
\caption{Google Research Football Hyperparameters}
\begin{tabular}{@{}llll@{}}
\toprule
Parameters                & Value                        & Parameters                & Value \\ \midrule
Critic LR         & 5e-4                          & Sample-to-insert Ratio    & 1.0   \\
Actor LR & 5e-4                          & Batch size                & 256  \\
Discount factor              & 0.995 & Number of mini-batch &  1    \\ 
Number of actors              & 256 & Max grad norm &  0.5    \\ 
Entropy coeff             & 1e-2 & Discount factor &  0.995    \\ 
Training epochs            & 10 & Rollout size &  1024    \\ 
Diffusion steps            & Num of players & Ratio clip  &  0.2    \\ 

\bottomrule
\end{tabular}
\end{table}

\subsection{Computation of the Stratified Bootstrap Confidence Interval}
To compute the stratified bootstrap confidence interval for aggregated Atari metrics, we use a resampling process designed to provide a statistically rigorous characterization of algorithm performance across high-variance environments. The computation involves performing 50,000 bootstrap iterations to ensure stability in the resulting estimates. In each iteration, a synthetic dataset of 168 runs is generated by \textbf{sampling with replacement} from the original pool of 3 seeds for each of the 56 tasks (in total 168 tasks). This "stratified" approach means that the resampling occurs independently within each environment (the strata), ensuring that each game contributes a consistent weight to the aggregate metric across every iteration. 

Specifically, let $X_{t,s}$ be the human-normalized score for task $t \in \{1, \dots, 56\}$ and seed $s \in \{1, \dots, 3\}$. The total number of runs is $M = 56 \times 3 = 168$.  In each of the $B = 50,000$ bootstrap iterations, we perform stratified resampling:  $$\mathcal{X}^*_b = \bigcup_{t=1}^{56} \{ X^*_{t,1}, \dots, X^*_{t,3} \}_b$$ where each $X^*_{t,s}$ is sampled with replacement from the original set of seeds $\{X_{t,1}, X_{t,2}, X_{t,3}\}$ for that specific task. This stratification ensures that every bootstrap sample $\mathcal{X}^*_b$ maintains exactly 3 runs per environment.  We then calculate the aggregate statistic, such as the Mean ($\mu^*_b$), Median, or Interquartile Mean (IQM):  $$\mu^*_b = \frac{1}{168} \sum_{i=1}^{168} x^*_i, \quad x^*_i \in \mathcal{X}^*_b$$ The IQM is calculated by discarding the lowest and highest 25\% of the $M$ scores in the resampled set and averaging the remaining 50\%. The final $95\%$ confidence interval is determined by the $2.5^{th}$ and $97.5^{th}$ percentiles of the empirical distribution formed by these $B$ bootstrap statistics:  $$[q_{0.025}(\text{stat}^*_1, \dots, \text{stat}^*_B), \quad q_{0.975}(\text{stat}^*_1, \dots, \text{stat}^*_B)]$$ This method allows the researchers to provide the robust intervals seen in Table 6 and Figure 8, which account for the high variance and non-normal distribution typical of reinforcement learning benchmarks.

For every one of the 50,000 resampled datasets, an aggregate statistic---such as the mean, median, or Interquartile Mean (IQM) of human-normalized scores---is calculated. The final 95\% confidence interval is derived from the distribution of these calculated statistics, offering a robust measure of uncertainty that, unlike standard error, does not rely on the assumption of a normal performance distribution.




\begin{table}[h]
    \centering
    \resizebox{\linewidth}{!}{
    \begin{tabular}{llllllll}
\toprule
Tasks & DQN & DQN-Macro & IMPALA & IMPALA-Macro & PPO & R2D2 & RL-D$^2$ \\
\midrule
Alien & 11.49k $\pm$ \scriptsize{105.46} & 2.63k $\pm$ \scriptsize{43.32} & 9.04k $\pm$ \scriptsize{201.70} & 12.40k $\pm$ \scriptsize{183.48} & 17.84k $\pm$ \scriptsize{480.29} & 8.09k $\pm$ \scriptsize{89.89} & 22.53k $\pm$ \scriptsize{848.31} \\
Amidar & 3.43k $\pm$ \scriptsize{37.04} & 691.56 $\pm$ \scriptsize{29.63} & 3.06k $\pm$ \scriptsize{12.48} & 5.60k $\pm$ \scriptsize{39.90} & 1.44k $\pm$ \scriptsize{1.89} & 1.64k $\pm$ \scriptsize{35.44} & 6.44k $\pm$ \scriptsize{71.30} \\
Assault & 5.92k $\pm$ \scriptsize{137.76} & 6.46k $\pm$ \scriptsize{433.42} & 18.34k $\pm$ \scriptsize{884.87} & 18.29k $\pm$ \scriptsize{467.93} & 5.23k $\pm$ \scriptsize{121.56} & 3.50k $\pm$ \scriptsize{96.09} & 20.99k $\pm$ \scriptsize{286.49} \\
Asterix & 4.56k $\pm$ \scriptsize{73.22} & 24.42k $\pm$ \scriptsize{611.26} & 361.80k $\pm$ \scriptsize{6.94k} & 28.07k $\pm$ \scriptsize{563.07} & 37.57k $\pm$ \scriptsize{1.92k} & 4.67k $\pm$ \scriptsize{64.12} & 133.56k $\pm$ \scriptsize{7.46k} \\
Asteroids & 1.51k $\pm$ \scriptsize{23.60} & 2.11k $\pm$ \scriptsize{15.32} & 5.71k $\pm$ \scriptsize{112.65} & 8.95k $\pm$ \scriptsize{145.30} & 14.49k $\pm$ \scriptsize{266.01} & 2.01k $\pm$ \scriptsize{32.60} & 82.24k $\pm$ \scriptsize{5.90k} \\
Atlantis & 984.11k $\pm$ \scriptsize{13.14k} & 802.66k $\pm$ \scriptsize{50.57k} & 1030.98k $\pm$ \scriptsize{6.81k} & 864.89k $\pm$ \scriptsize{1.24k} & 710.85k $\pm$ \scriptsize{18.11k} & 1082.26k $\pm$ \scriptsize{40.63k} & 1003.63k $\pm$ \scriptsize{1.79k} \\
BankHeist & 1.84k $\pm$ \scriptsize{18.02} & 1.25k $\pm$ \scriptsize{19.15} & 1.50k $\pm$ \scriptsize{2.70} & 1.10k $\pm$ \scriptsize{4.31} & 485.20 $\pm$ \scriptsize{4.21} & 944.20 $\pm$ \scriptsize{3.53} & 1.70k $\pm$ \scriptsize{6.59} \\
BattleZone & 116.15k $\pm$ \scriptsize{1.48k} & 43.66k $\pm$ \scriptsize{837.10} & 68.43k $\pm$ \scriptsize{1.12k} & 166.30k $\pm$ \scriptsize{2.96k} & 54.66k $\pm$ \scriptsize{472.35} & 76.09k $\pm$ \scriptsize{1.69k} & 197.94k $\pm$ \scriptsize{3.08k} \\
BeamRider & 3.55k $\pm$ \scriptsize{99.35} & 5.62k $\pm$ \scriptsize{815.84} & 21.07k $\pm$ \scriptsize{837.41} & 14.82k $\pm$ \scriptsize{143.88} & 29.47k $\pm$ \scriptsize{774.52} & 3.10k $\pm$ \scriptsize{78.03} & 28.98k $\pm$ \scriptsize{1.10k} \\
Berzerk & 340.95k $\pm$ \scriptsize{24.42k} & 1.11k $\pm$ \scriptsize{19.22} & 1.49k $\pm$ \scriptsize{45.20} & 7.43k $\pm$ \scriptsize{269.77} & 1.29k $\pm$ \scriptsize{33.72} & 110.48k $\pm$ \scriptsize{42.94} & 802.99k $\pm$ \scriptsize{64.87k} \\
Bowling & 266.38 $\pm$ \scriptsize{0.01} & 41.94 $\pm$ \scriptsize{6.70} & 70.00 $\pm$ \scriptsize{0.00} & 54.80 $\pm$ \scriptsize{0.13} & 149.03 $\pm$ \scriptsize{0.25} & 197.18 $\pm$ \scriptsize{0.17} & 266.38 $\pm$ \scriptsize{0.00} \\
Boxing & 97.13 $\pm$ \scriptsize{0.20} & 99.31 $\pm$ \scriptsize{0.10} & 100.00 $\pm$ \scriptsize{0.00} & 98.60 $\pm$ \scriptsize{0.02} & 98.99 $\pm$ \scriptsize{0.11} & 97.89 $\pm$ \scriptsize{0.12} & 99.51 $\pm$ \scriptsize{0.00} \\
Breakout & 76.39 $\pm$ \scriptsize{2.94} & 401.10 $\pm$ \scriptsize{1.61} & 675.48 $\pm$ \scriptsize{18.21} & 161.19 $\pm$ \scriptsize{5.48} & 394.17 $\pm$ \scriptsize{0.69} & 124.08 $\pm$ \scriptsize{1.44} & 424.15 $\pm$ \scriptsize{0.00} \\
Centipede & 36.42k $\pm$ \scriptsize{541.25} & 6.26k $\pm$ \scriptsize{214.74} & 8.08k $\pm$ \scriptsize{381.00} & 27.66k $\pm$ \scriptsize{835.60} & 27.60k $\pm$ \scriptsize{380.41} & 20.66k $\pm$ \scriptsize{247.27} & 68.41k $\pm$ \scriptsize{910.77} \\
ChopperCommand & 13.18k $\pm$ \scriptsize{320.75} & 3.59k $\pm$ \scriptsize{300.73} & 23.86k $\pm$ \scriptsize{630.67} & 15.74k $\pm$ \scriptsize{709.66} & 2.35k $\pm$ \scriptsize{115.21} & 2.52k $\pm$ \scriptsize{148.92} & 34.36k $\pm$ \scriptsize{609.01} \\
CrazyClimber & 93.75k $\pm$ \scriptsize{1.37k} & 136.37k $\pm$ \scriptsize{3.07k} & 136.05k $\pm$ \scriptsize{949.62} & 107.05k $\pm$ \scriptsize{1.55k} & 70.01k $\pm$ \scriptsize{1.19k} & 118.39k $\pm$ \scriptsize{1.04k} & 113.69k $\pm$ \scriptsize{759.99} \\
Defender & 17.60k $\pm$ \scriptsize{256.97} & 51.94k $\pm$ \scriptsize{496.01} & 427.49k $\pm$ \scriptsize{23.25k} & 33.85k $\pm$ \scriptsize{618.50} & 55.75k $\pm$ \scriptsize{606.50} & 38.70k $\pm$ \scriptsize{961.61} & 138.16k $\pm$ \scriptsize{7.10k} \\
DemonAttack & 3.89k $\pm$ \scriptsize{61.93} & 26.78k $\pm$ \scriptsize{5.85k} & 132.40k $\pm$ \scriptsize{126.41} & 44.64k $\pm$ \scriptsize{1.58k} & 9.91k $\pm$ \scriptsize{313.53} & 2.89k $\pm$ \scriptsize{35.35} & 55.93k $\pm$ \scriptsize{1.34k} \\
DoubleDunk & -0.36 $\pm$ \scriptsize{0.06} & -3.18 $\pm$ \scriptsize{2.10} & 23.46 $\pm$ \scriptsize{0.07} & 0.00 $\pm$ \scriptsize{0.00} & 5.60 $\pm$ \scriptsize{0.36} & -0.86 $\pm$ \scriptsize{0.34} & 24.00 $\pm$ \scriptsize{0.00} \\
Enduro & 651.59 $\pm$ \scriptsize{8.84} & 2.22k $\pm$ \scriptsize{58.92} & 8.16 $\pm$ \scriptsize{0.49} & 1.15k $\pm$ \scriptsize{19.29} & 1.64k $\pm$ \scriptsize{48.10} & 852.35 $\pm$ \scriptsize{29.13} & 1.45k $\pm$ \scriptsize{50.17} \\
FishingDerby & 68.54 $\pm$ \scriptsize{0.86} & 26.46 $\pm$ \scriptsize{0.52} & 44.99 $\pm$ \scriptsize{0.66} & 45.31 $\pm$ \scriptsize{0.68} & 0.20 $\pm$ \scriptsize{0.67} & 20.74 $\pm$ \scriptsize{0.99} & 80.64 $\pm$ \scriptsize{0.00} \\
Freeway & 33.82 $\pm$ \scriptsize{0.00} & 32.61 $\pm$ \scriptsize{0.07} & 32.73 $\pm$ \scriptsize{0.03} & 33.46 $\pm$ \scriptsize{0.03} & 32.68 $\pm$ \scriptsize{0.06} & 34.00 $\pm$ \scriptsize{0.00} & 33.84 $\pm$ \scriptsize{0.00} \\
Frostbite & 9.41k $\pm$ \scriptsize{14.63} & 3.93k $\pm$ \scriptsize{194.22} & 862.00 $\pm$ \scriptsize{48.06} & 9.00k $\pm$ \scriptsize{2.64} & 3.97k $\pm$ \scriptsize{429.59} & 10.47k $\pm$ \scriptsize{78.34} & 14.39k $\pm$ \scriptsize{143.25} \\
Gopher & 2.72k $\pm$ \scriptsize{120.47} & 27.47k $\pm$ \scriptsize{3.09k} & 89.58k $\pm$ \scriptsize{3.91k} & 27.69k $\pm$ \scriptsize{626.48} & 5.38k $\pm$ \scriptsize{132.20} & 3.95k $\pm$ \scriptsize{274.47} & 67.92k $\pm$ \scriptsize{1.32k} \\
Gravitar & 2.68k $\pm$ \scriptsize{11.18} & 1.10k $\pm$ \scriptsize{19.89} & 4.27k $\pm$ \scriptsize{2.64} & 4.87k $\pm$ \scriptsize{28.31} & 4.57k $\pm$ \scriptsize{5.56} & 3.08k $\pm$ \scriptsize{28.87} & 4.62k $\pm$ \scriptsize{13.83} \\
Hero & 22.90k $\pm$ \scriptsize{17.41} & 10.69k $\pm$ \scriptsize{452.97} & 28.98k $\pm$ \scriptsize{5.71} & 36.74k $\pm$ \scriptsize{7.49} & 14.08k $\pm$ \scriptsize{10.82} & 32.41k $\pm$ \scriptsize{685.43} & 28.94k $\pm$ \scriptsize{14.34} \\
IceHockey & 12.45 $\pm$ \scriptsize{0.70} & 4.62 $\pm$ \scriptsize{0.67} & 26.42 $\pm$ \scriptsize{0.15} & 25.69 $\pm$ \scriptsize{0.20} & 15.98 $\pm$ \scriptsize{0.45} & -0.15 $\pm$ \scriptsize{0.16} & 46.60 $\pm$ \scriptsize{0.14} \\
Jamesbond & 1.41k $\pm$ \scriptsize{50.94} & 584.00 $\pm$ \scriptsize{9.20} & 1.74k $\pm$ \scriptsize{66.32} & 75.06k $\pm$ \scriptsize{80.92} & 1.60k $\pm$ \scriptsize{155.90} & 1.08k $\pm$ \scriptsize{18.32} & 18.21k $\pm$ \scriptsize{3.59k} \\
Kangaroo & 14.67k $\pm$ \scriptsize{113.45} & 8.96k $\pm$ \scriptsize{273.42} & 14.50k $\pm$ \scriptsize{0.00} & 14.22k $\pm$ \scriptsize{8.61} & 2.00k $\pm$ \scriptsize{0.00} & 12.80k $\pm$ \scriptsize{127.16} & 15.26k $\pm$ \scriptsize{10.64} \\
Krull & 69.80k $\pm$ \scriptsize{1.86k} & 9.36k $\pm$ \scriptsize{132.50} & 10.02k $\pm$ \scriptsize{38.49} & 83.59k $\pm$ \scriptsize{2.56k} & 8.97k $\pm$ \scriptsize{83.18} & 61.42k $\pm$ \scriptsize{1.50k} & 385.51k $\pm$ \scriptsize{3.70k} \\
KungFuMaster & 26.71k $\pm$ \scriptsize{236.87} & 36.65k $\pm$ \scriptsize{718.07} & 55.69k $\pm$ \scriptsize{1.09k} & 15.98k $\pm$ \scriptsize{206.24} & 31.15k $\pm$ \scriptsize{465.11} & 54.57k $\pm$ \scriptsize{1.02k} & 63.76k $\pm$ \scriptsize{789.60} \\
MontezumaRevenge & 835.28 $\pm$ \scriptsize{36.72} & 0.00 $\pm$ \scriptsize{0.00} & 0.00 $\pm$ \scriptsize{0.00} & 0.00 $\pm$ \scriptsize{0.00} & 400.00 $\pm$ \scriptsize{0.00} & 400.00 $\pm$ \scriptsize{0.00} & 2.50k $\pm$ \scriptsize{7.26} \\
MsPacman & 20.18k $\pm$ \scriptsize{162.54} & 3.90k $\pm$ \scriptsize{130.60} & 8.85k $\pm$ \scriptsize{77.01} & 7.77k $\pm$ \scriptsize{63.70} & 21.04k $\pm$ \scriptsize{250.98} & 9.82k $\pm$ \scriptsize{68.46} & 22.87k $\pm$ \scriptsize{3.39} \\
NameThisGame & 6.86k $\pm$ \scriptsize{68.39} & 14.91k $\pm$ \scriptsize{442.47} & 15.64k $\pm$ \scriptsize{90.75} & 13.69k $\pm$ \scriptsize{109.46} & 8.40k $\pm$ \scriptsize{68.86} & 7.15k $\pm$ \scriptsize{81.37} & 12.66k $\pm$ \scriptsize{197.84} \\
Phoenix & 4.98k $\pm$ \scriptsize{66.88} & 14.11k $\pm$ \scriptsize{1.41k} & 192.34k $\pm$ \scriptsize{11.13k} & 6.37k $\pm$ \scriptsize{38.35} & 43.91k $\pm$ \scriptsize{3.04k} & 5.61k $\pm$ \scriptsize{46.40} & 202.39k $\pm$ \scriptsize{5.33k} \\
Pitfall & 0.00 $\pm$ \scriptsize{0.00} & -2.48 $\pm$ \scriptsize{0.94} & 0.00 $\pm$ \scriptsize{0.00} & 0.00 $\pm$ \scriptsize{0.00} & 0.00 $\pm$ \scriptsize{0.00} & 0.00 $\pm$ \scriptsize{0.00} & 0.00 $\pm$ \scriptsize{0.00} \\
Pong & 13.31 $\pm$ \scriptsize{0.12} & 19.70 $\pm$ \scriptsize{0.13} & 20.47 $\pm$ \scriptsize{0.09} & 12.25 $\pm$ \scriptsize{0.22} & 21.00 $\pm$ \scriptsize{0.00} & 19.73 $\pm$ \scriptsize{0.11} & 21.00 $\pm$ \scriptsize{0.00} \\
PrivateEye & 35.22k $\pm$ \scriptsize{4.16} & 100.00 $\pm$ \scriptsize{0.00} & 389.34 $\pm$ \scriptsize{268.73} & 99.64 $\pm$ \scriptsize{0.02} & 1.59k $\pm$ \scriptsize{0.57} & 35.12k $\pm$ \scriptsize{28.48} & 15.18k $\pm$ \scriptsize{4.02} \\
Qbert & 29.68k $\pm$ \scriptsize{62.59} & 9.00k $\pm$ \scriptsize{431.38} & 15.66k $\pm$ \scriptsize{386.65} & 475.04 $\pm$ \scriptsize{0.03} & 5.49k $\pm$ \scriptsize{263.05} & 16.68k $\pm$ \scriptsize{213.54} & 30.85k $\pm$ \scriptsize{57.30} \\
Riverraid & 7.48k $\pm$ \scriptsize{81.04} & 17.07k $\pm$ \scriptsize{492.67} & 18.44k $\pm$ \scriptsize{100.97} & 9.87k $\pm$ \scriptsize{39.11} & 8.01k $\pm$ \scriptsize{56.74} & 11.33k $\pm$ \scriptsize{90.61} & 17.94k $\pm$ \scriptsize{98.91} \\
RoadRunner & 317.32k $\pm$ \scriptsize{9.62k} & 52.61k $\pm$ \scriptsize{630.59} & 59.20k $\pm$ \scriptsize{368.83} & 514.91k $\pm$ \scriptsize{7.59k} & 23.88k $\pm$ \scriptsize{212.10} & 77.74k $\pm$ \scriptsize{1.95k} & 563.86k $\pm$ \scriptsize{2.53k} \\
Robotank & 33.33 $\pm$ \scriptsize{0.72} & 66.72 $\pm$ \scriptsize{0.49} & 71.97 $\pm$ \scriptsize{0.33} & 66.67 $\pm$ \scriptsize{0.17} & 63.62 $\pm$ \scriptsize{0.51} & 31.22 $\pm$ \scriptsize{0.20} & 66.74 $\pm$ \scriptsize{0.79} \\
Seaquest & 3.97k $\pm$ \scriptsize{37.56} & 24.67k $\pm$ \scriptsize{5.63k} & 27.12k $\pm$ \scriptsize{824.83} & 10.85k $\pm$ \scriptsize{134.19} & 6.87k $\pm$ \scriptsize{72.78} & 3.63k $\pm$ \scriptsize{194.09} & 144.62k $\pm$ \scriptsize{4.65k} \\
Skiing & -4.43k $\pm$ \scriptsize{5.26} & -29.41k $\pm$ \scriptsize{266.48} & -9.01k $\pm$ \scriptsize{0.07} & -8.95k $\pm$ \scriptsize{0.00} & -15.00k $\pm$ \scriptsize{111.88} & -27.93k $\pm$ \scriptsize{185.34} & -4.41k $\pm$ \scriptsize{7.47} \\
Solaris & 15.00k $\pm$ \scriptsize{517.51} & 3.26k $\pm$ \scriptsize{181.94} & 2.22k $\pm$ \scriptsize{113.16} & 2.41k $\pm$ \scriptsize{47.98} & 6.00k $\pm$ \scriptsize{146.73} & 3.02k $\pm$ \scriptsize{160.32} & 14.39k $\pm$ \scriptsize{263.47} \\
SpaceInvaders & 2.02k $\pm$ \scriptsize{30.80} & 6.22k $\pm$ \scriptsize{714.13} & 41.54k $\pm$ \scriptsize{8.75k} & 9.92k $\pm$ \scriptsize{329.91} & 3.87k $\pm$ \scriptsize{56.20} & 1.75k $\pm$ \scriptsize{59.99} & 10.86k $\pm$ \scriptsize{480.94} \\
StarGunner & 1.40k $\pm$ \scriptsize{22.05} & 96.97k $\pm$ \scriptsize{7.53k} & 142.15k $\pm$ \scriptsize{838.43} & 29.10k $\pm$ \scriptsize{850.92} & 33.55k $\pm$ \scriptsize{427.98} & 2.23k $\pm$ \scriptsize{42.72} & 52.69k $\pm$ \scriptsize{448.91} \\
Tennis & 0.00 $\pm$ \scriptsize{0.00} & 20.77 $\pm$ \scriptsize{0.54} & 0.00 $\pm$ \scriptsize{0.00} & 0.00 $\pm$ \scriptsize{0.00} & -2.34 $\pm$ \scriptsize{0.58} & -1.69 $\pm$ \scriptsize{0.51} & 22.20 $\pm$ \scriptsize{0.17} \\
TimePilot & 34.33k $\pm$ \scriptsize{404.16} & 10.66k $\pm$ \scriptsize{159.67} & 55.38k $\pm$ \scriptsize{918.01} & 109.15k $\pm$ \scriptsize{4.05k} & 33.82k $\pm$ \scriptsize{544.74} & 11.37k $\pm$ \scriptsize{222.24} & 42.26k $\pm$ \scriptsize{722.78} \\
Tutankham & 160.03 $\pm$ \scriptsize{1.36} & 213.78 $\pm$ \scriptsize{8.55} & 240.49 $\pm$ \scriptsize{9.66} & 187.71 $\pm$ \scriptsize{0.03} & 190.20 $\pm$ \scriptsize{1.81} & 120.27 $\pm$ \scriptsize{3.63} & 230.73 $\pm$ \scriptsize{0.84} \\
UpNDown & 76.44k $\pm$ \scriptsize{910.55} & 65.52k $\pm$ \scriptsize{9.48k} & 422.67k $\pm$ \scriptsize{2.82k} & 322.33k $\pm$ \scriptsize{3.00k} & 240.21k $\pm$ \scriptsize{9.79k} & 127.06k $\pm$ \scriptsize{2.61k} & 264.48k $\pm$ \scriptsize{2.48k} \\
Venture & 2.05k $\pm$ \scriptsize{12.84} & 1.50k $\pm$ \scriptsize{35.34} & 20.00 $\pm$ \scriptsize{0.00} & 1.99k $\pm$ \scriptsize{2.95} & 1.41k $\pm$ \scriptsize{24.61} & 1.50k $\pm$ \scriptsize{24.88} & 2.06k $\pm$ \scriptsize{3.76} \\
VideoPinball & 155.83k $\pm$ \scriptsize{3.47k} & 341.15k $\pm$ \scriptsize{59.03k} & 542.75k $\pm$ \scriptsize{15.87k} & 392.46k $\pm$ \scriptsize{17.02k} & 63.37k $\pm$ \scriptsize{4.76k} & 117.67k $\pm$ \scriptsize{3.64k} & 545.38k $\pm$ \scriptsize{76.91k} \\
WizardOfWor & 31.94k $\pm$ \scriptsize{1.13k} & 16.52k $\pm$ \scriptsize{1.16k} & 19.41k $\pm$ \scriptsize{371.79} & 39.40k $\pm$ \scriptsize{406.34} & 9.68k $\pm$ \scriptsize{1.12k} & 10.94k $\pm$ \scriptsize{529.67} & 58.41k $\pm$ \scriptsize{588.34} \\
YarsRevenge & 82.41k $\pm$ \scriptsize{1.41k} & 70.73k $\pm$ \scriptsize{1.80k} & 125.27k $\pm$ \scriptsize{1.07k} & 139.64k $\pm$ \scriptsize{1.55k} & 70.65k $\pm$ \scriptsize{600.98} & 75.95k $\pm$ \scriptsize{785.33} & 165.95k $\pm$ \scriptsize{245.07} \\
Zaxxon & 18.84k $\pm$ \scriptsize{311.75} & 12.58k $\pm$ \scriptsize{727.69} & 34.99k $\pm$ \scriptsize{233.45} & 45.80k $\pm$ \scriptsize{276.10} & 30.07k $\pm$ \scriptsize{262.07} & 13.28k $\pm$ \scriptsize{505.46} & 30.60k $\pm$ \scriptsize{310.47} \\
\bottomrule
\end{tabular}
    }
    \caption{Full Performance of Atari Games.}
    \label{tab:full}
\end{table}

\section{Additional Experimental Results}
\label{sec.appendix.additional_exp}

\subsection{More Atari Aggregated Results}

\begin{figure}[H]
    \centering
    \includegraphics[width=0.9\linewidth]{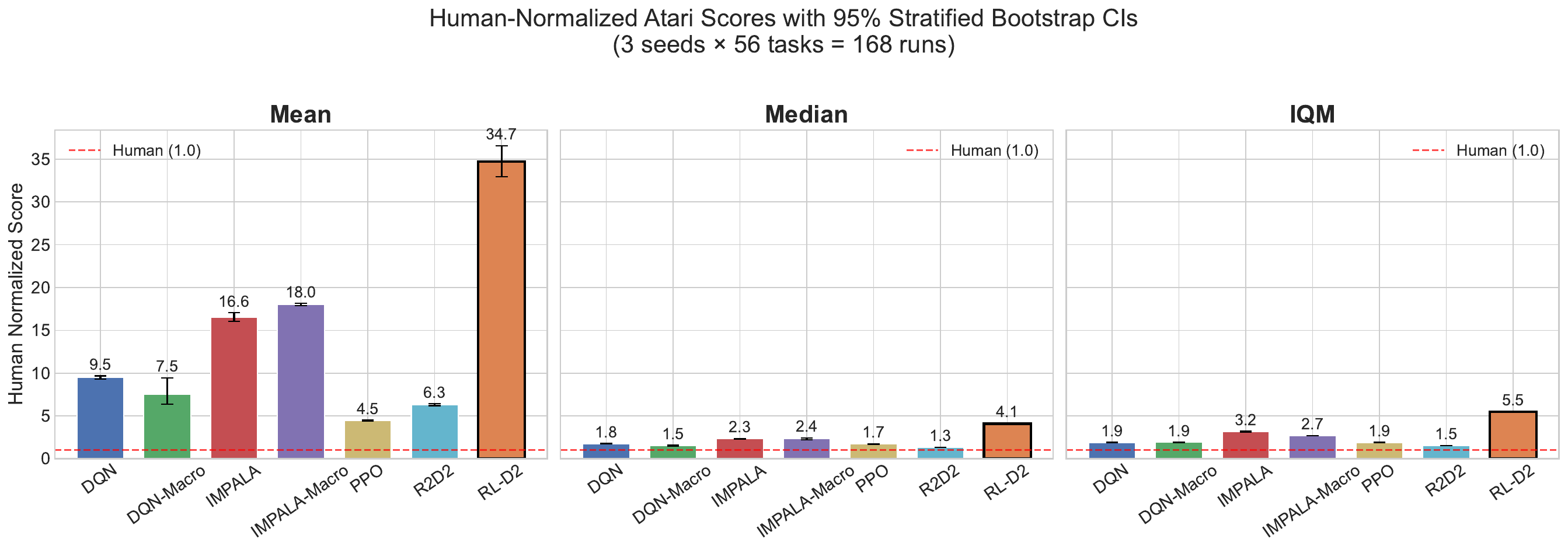}
    \caption{Statistics of Human-Normalized Atari scores. 95\% Confidence intervals are computed with stratified bootstrap sampling of 50,000 bootstrap iterations from 168 runs (3 seeds * 56 tasks).}
    \label{fig:atari_aggregate_statistics}
\end{figure}

\begin{table}[H]
\caption{Detailed numbers of the statistics and confidence intervals. Confidence intervals are computed with stratified bootstrap sampling of 50,000 bootstrap iterations.}
\centering
\begin{tabular}{lccc}
\hline
 & \textbf{Mean} & \textbf{Median} & \textbf{IQM} \\
\hline
DQN & 9.49 [ 9.31, 9.68] & 1.77 [ 1.76, 1.81] & 1.89 [ 1.88, 1.90] \\
DQN-Macro & 7.54 [ 6.38, 9.46] & 1.51 [ 1.44, 1.57] & 1.91 [ 1.88, 1.94] \\
IMPALA & 16.55 [16.07, 17.07] & 2.33 [ 2.30, 2.37] & 3.17 [ 3.08, 3.23] \\
IMPALA-Macro & 18.01 [17.88, 18.15] & 2.36 [ 2.22, 2.45] & 2.69 [ 2.66, 2.71] \\
PPO & 4.47 [ 4.39, 4.54] & 1.73 [ 1.68, 1.76] & 1.90 [ 1.88, 1.91] \\
R2D2 & 6.31 [ 6.20, 6.41] & 1.32 [ 1.32, 1.34] & 1.55 [ 1.54, 1.56] \\
RL-D$^2$ & \textbf{31.33 [29.74, 33.00]} & \textbf{4.09 [ 4.00,  4.20] } & \textbf{4.98 [ 4.96, 5.00]} \\
\hline
\end{tabular}
\end{table}

\begin{figure}[h]
    \centering
    \includegraphics[width=\linewidth]{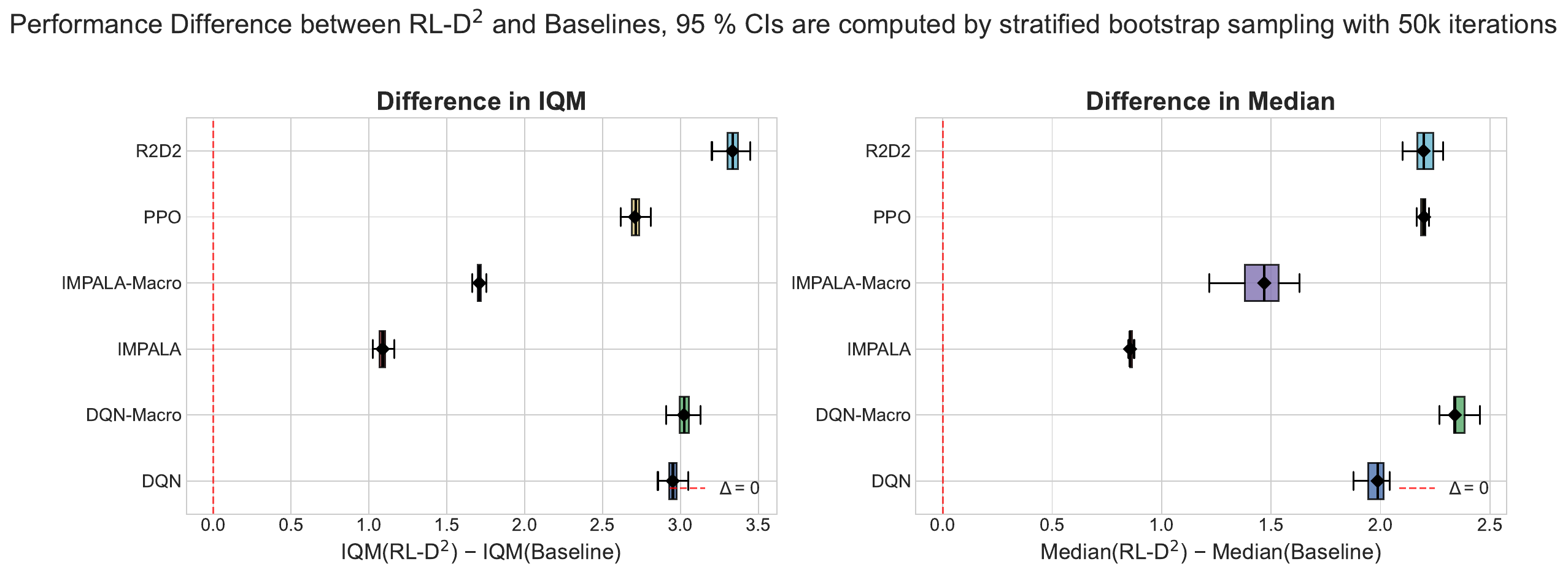}
    \caption{Performance difference statistics. 95\% CIs are computed by stratified bootstrap sampling with 50k iterations from 168 runs (3 seeds * 56 tasks).}
    \label{fig:atari_performance_difference}
\end{figure}

\subsection{Full Results for Atari Games}
\label{sec.full_results_atari}
\begin{figure}[h]
    \centering
    \includegraphics[width=\linewidth]{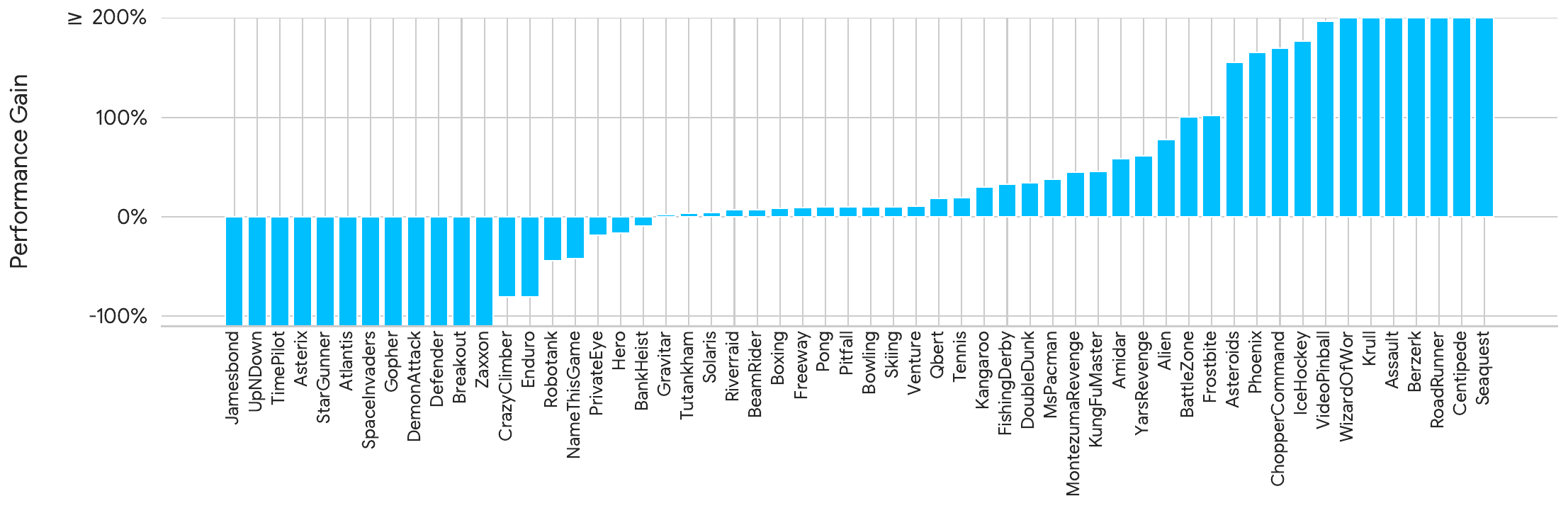}
    \caption{Mean human normalized score of RL-D$^2$ compared to the best baselines in each Atari task.}
    \label{fig:big_bar}
\end{figure}

Please refer to~\Cref{tab:full} for the full results of Atari games, and \Cref{fig:big_bar} for the comparison with the best baselines. We outperform all baselines in 36 out of 56 Atari environments.

\subsection{Ablation of temperature tuning} 
\label{sec.appendix.temperature}

{We conduct ablation studies on the temperature tuning discussed in \Cref{sec.appendix.temp_tuning} on multiple benchmarks including MinAtar, Atari and Google Football. }

 \subsubsection{MinAtar}

\begin{figure}[h]
    \centering
    \includegraphics[width=\linewidth]{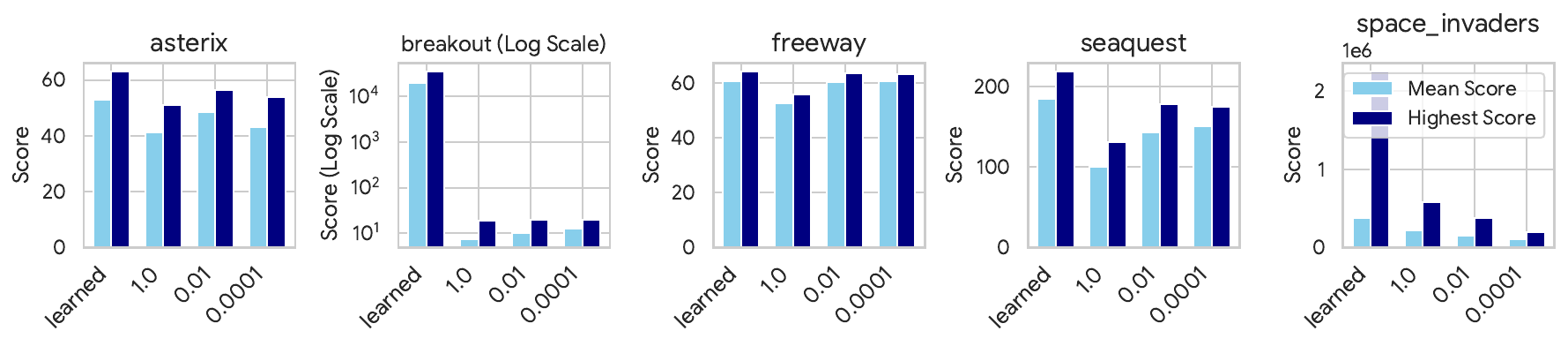}
    \caption{Ablation studies of temperature tuning {with fixed temperature variable}. Bars indicate the mean episode returns over the last 100 evaluations over 3 seeds. }
    \label{fig:appendix_temp}
\end{figure}

 For MinAtar, the current temperature is updated following a KL constraint is set to 1.0. We compare auto-tuning with fixed temperatures. The results are shown in~\Cref{fig:appendix_temp}, 
 showing the auto-tuning consistently outperforms fixed temperature. 

\begin{figure}[h]
    \centering
    \includegraphics[width=0.8\linewidth]{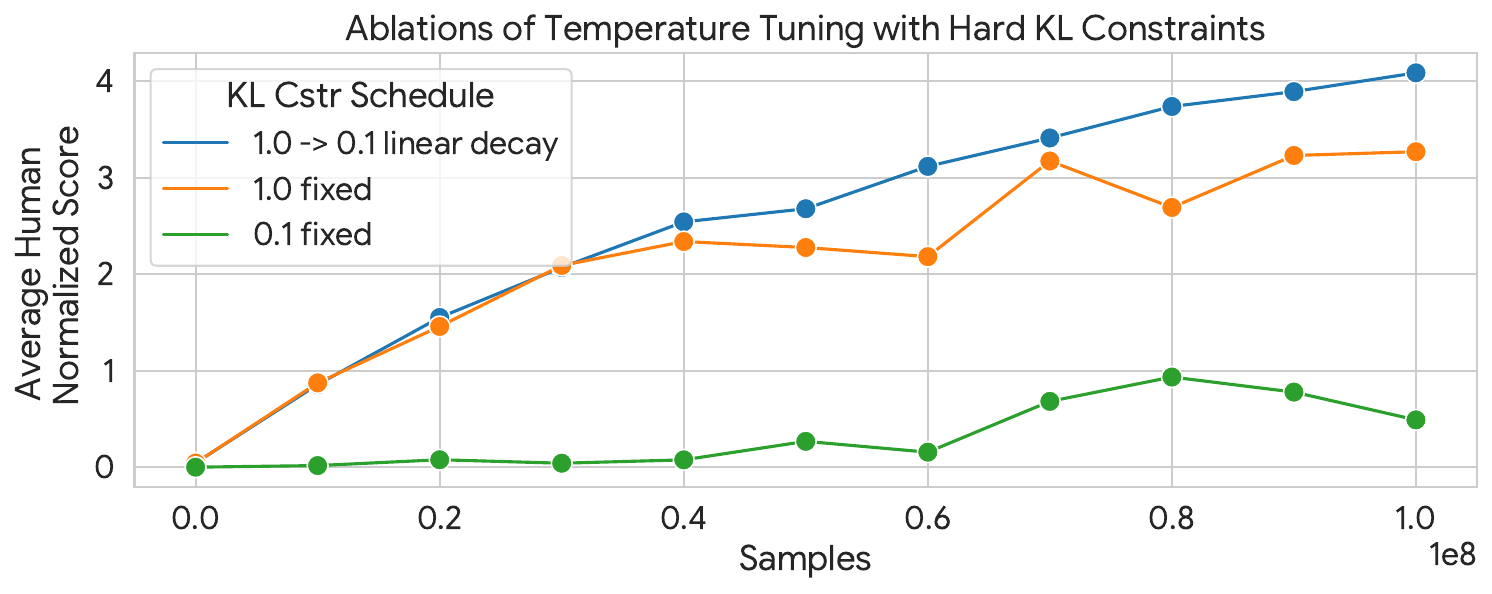}
    \caption{Ablation studies of temperature tuning with hard KL constraints, compared with different KL constraint schedules. The results show the average human-normalized score over 10 Atari environments, each with 3 random seeds.}
    \label{fig:appendix_temp_atari}
\end{figure}

 {
 \subsubsection{Atari Games} For Atari games, we use KL constraint schedule linearly decaying from 1 to 0.1 in the first $10^7$ samples. The intuition behind this selection is that, in vast combinatorial discrete spaces, a larger initial KL constraint allows the policy to deviate significantly from initialization, enabling the broad exploration necessary to discover high-value actions.

 We compare auto-tuning with {two other temperature control schemes: fixed KL constraints 1.0 and 0.1.} The results are shown in~\Cref{fig:appendix_temp_atari}, showing that smaller KL constraints like 0.1 fail to learn in the initial phase. Large KL constraints like 1.0 cause instabilities after 50M steps, which makes the performance worse. The linear decay achieves initial fast learning and overall stability. 
}

\begin{figure}[h]
    \centering
    \includegraphics[width=0.8\linewidth]{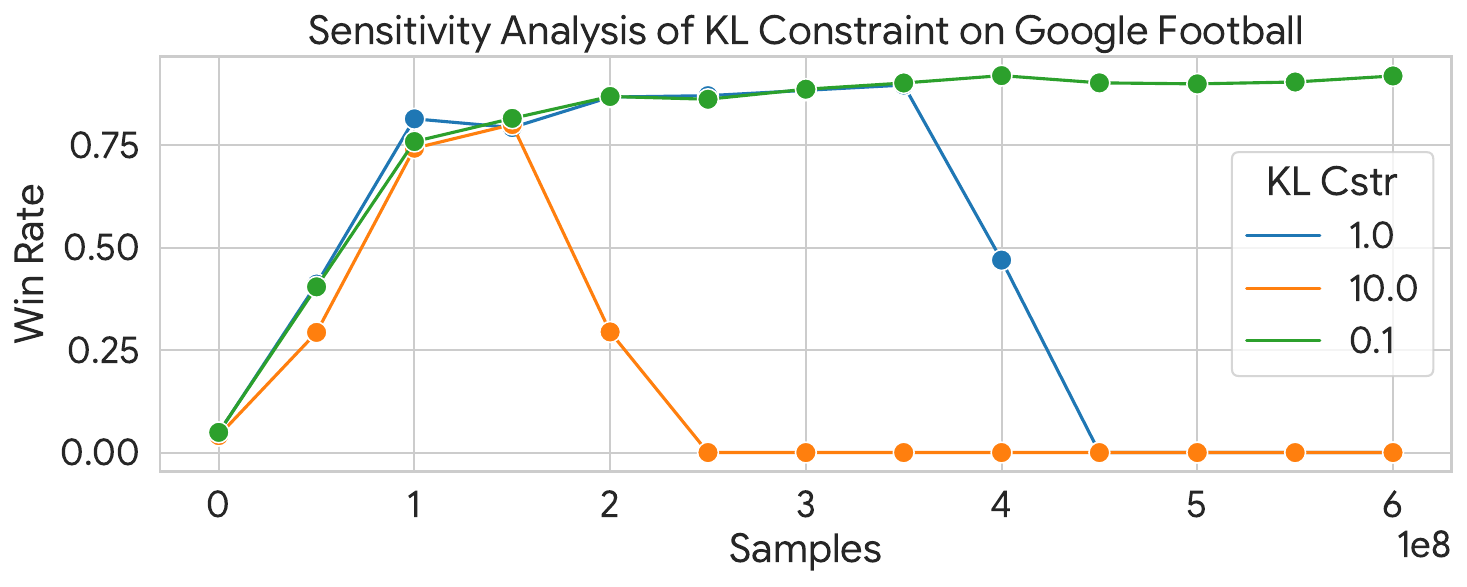}
    \caption{Sensitivity studies of temperature tuning with hard KL constraints on Google Football 11 vs 11. The results show the average win rate over 5 random seeds.}
    \label{fig:appendix_temp_gft}
\end{figure}

{
\subsubsection{Google Football}
\label{sec.appendix.temp.gft}
We compare three different KL constraint schedules in Google Football: 0.1, 1.0, and 10. The results are shown in \Cref{fig:appendix_temp_gft}, which shows another failure mode of collapsing for larger KL constraints. The reason might be that larger KL constraints push the policy to be highly deterministic. Therefore, the collected data loses diversity, making training fail.
}

{
In summary, temperature tuning is significant for the performance of RL-D$^2$-FKL. When selecting the KL constraints, we should consider two key factors: (1) The initial KL constraint should be large so that the policy can start learning; (2) collapse should be avoided in the later training phase by enforcing KL constraints that are not too large. Although not necessary, a decay schedule is very helpful for stabilizing training and improving performance.
}

\subsection{Discrete Diffusion as Planner for Causal Action Spaces} 

In applications of macro actions in Atari games, we can commit to only the first action rather than all the macro actions. Therefore, it is common to plan for a longer trajectory and only commit to the first action, such as model predictive control and Monte-Carlo tree search~\citep{garcia1989model,silver2016mastering}. However, if we would like to implement planning in online RL, the parameterization of the planning trajectory is not trivial. If we use autoregressive models as the parameterization to generate the trajectory, the next action will depend only on the current state and not on the planned trajectory, leaving future actions useless. 

\begin{figure}
    \centering
    \includegraphics[width=0.9\linewidth]{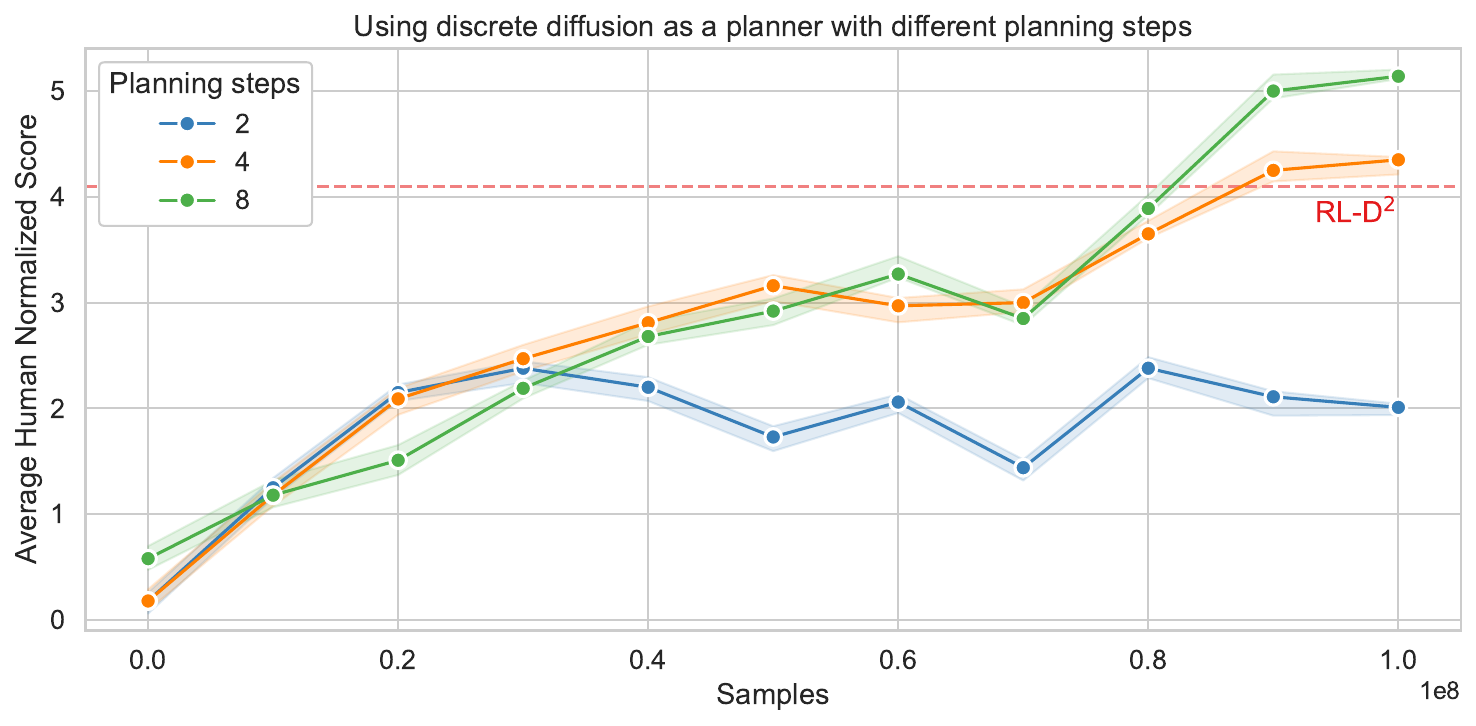}
    \caption{Performance using discrete diffusion as a planner with different planning-step lengths, averaged over 10 Atari games with 3 random seeds for each game.}
    \label{fig:planner}
\end{figure}

Benefiting from the non-causal unmasking of discrete diffusion models, we can directly use discrete diffusion to parameterize the planner. \textbf{Note that this is a totally different setup and algorithm from the main text \Cref{sec:main_exp}}. {We show the planner performance with the same set of hyperparameters in \Cref{tab:atari_hyperparameters} with varying planning steps shown in~\Cref{fig:planner}. The performance increases with increasing planning length, showing that committing only to the first action is more robust than committing to all macro actions. The planner is also scalable with respect to the increasing size of action spaces.}

\begin{algorithm}
    \caption{Discrete Diffusion as Planner}\label{alg.planning}
    \begin{algorithmic}[1]
        \REQUIRE Planning length $K$, current policy $\pi_{\rm old}$, replay buffer $\mathcal{D}$, current value function $q^{\pi_{\rm old}}$.
        \STATE \texttt{\textcolor{blue}{\# Policy updates in training.}}
        \STATE For states $s\sim\mathcal{D}$, sample macro actions $\textbf{a}=(a_1,\dots,a_k)$, compute $\gL_{\rm ELBO}$ with~\Cref{eq:elbo}.
        \STATE Take the first one to compute value function $q^{\pi_{\rm old}}(s, a_1)$.
        \STATE Optimize the policy by self-imitation loss~\Cref{eq:fkl_loss}.

        \STATE \texttt{\textcolor{blue}{\# Inference.}}
        \STATE Sample macro actions $(a_1,\dots,a_k)$ and only take $a=a_0$.
    \end{algorithmic}
\end{algorithm}

\subsection{Ablation of On-Policy Diffusion Training.}
\begin{figure}[h]
    \centering
    \includegraphics[width=0.8\linewidth]{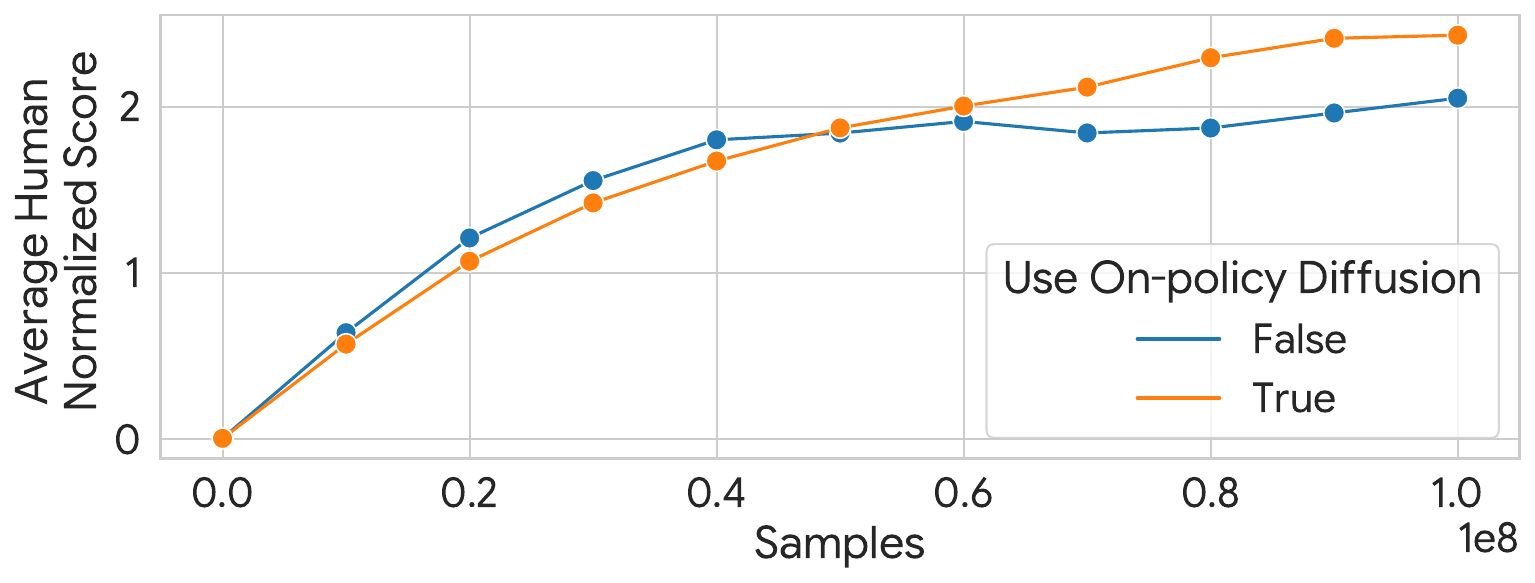}
    \caption{Ablation studies of on-policy diffusion training. The curves indicate mean reward using macro length 8 over 10 representative Atari environments, 3 seeds, and 10 consecutive evaluation episodes.}
    \label{fig:ablation_onpolicy}
\end{figure}

We conduct ablation studies on the on-policy training discussed in \Cref{sec.onpolicy_training}, and the results are shown in \Cref{fig:ablation_onpolicy}, which shows that on-policy diffusion training helps improve performance.

\subsection{Ablation of Network Architecture.}

We compare using multi-layer perceptrons (MLP) versus transformers~\citep{vaswani2017attention} for parameterization of the $Q$ and policy networks. The two networks share the same number of parameters, around $4\times 10^5$. We can see that the transformer consistently outperforms MLP.

\begin{figure}[h]
    \centering
    \includegraphics[width=\linewidth]{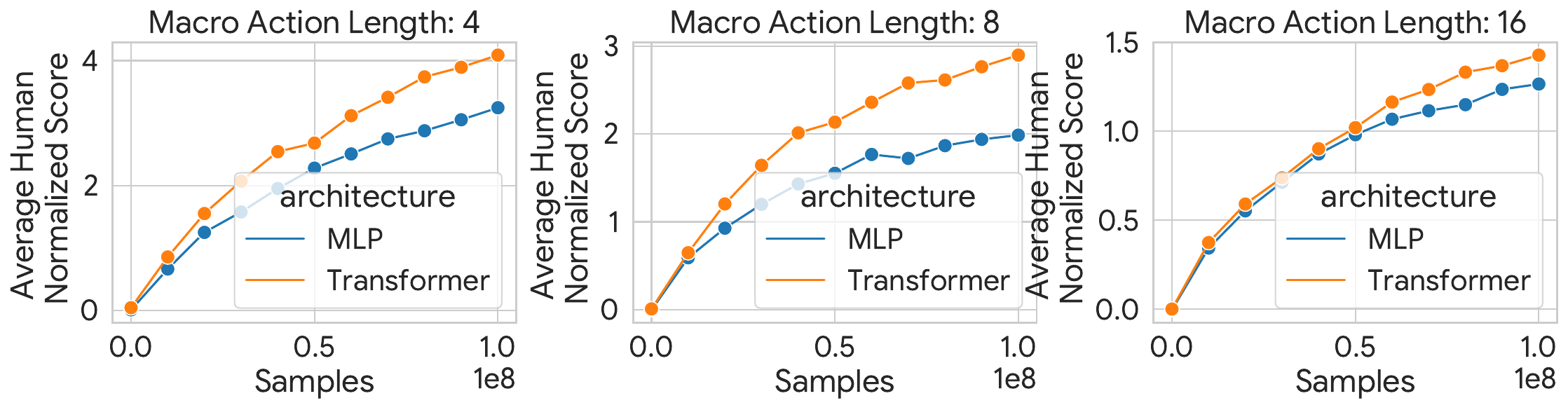}
    \caption{Ablation studies of network architecture. The curves indicate mean reward using macro length 8 over 10 representative Atari environments, 3 seeds, and 10 consecutive evaluation episodes.}
    \label{fig:on_policy_ablation}
\end{figure}

\section{Baseline Selection Protocol}

Our baseline selection protocol was designed to rigorously evaluate the performance, scalability, and versatility of \textbf{$RL-D^{2}$} across the distinct challenges presented by combinatorial action spaces. We selected a diverse set of baselines, ranging from established standards in reinforcement learning to domain-specific state-of-the-art methods, ensuring that our comparisons were fair, comprehensive, and directly addressed the central claims of our work.

The selection was tailored to the three specific experimental domains:

\subsection{DNA Sequence Generation}
This domain tests the single-step policy optimization (combinatorial bandit) capabilities of our framework. The baselines were chosen to cover two main categories:

\begin{itemize}
    \item \textbf{Controlled Generation Methods:} We included standard guidance-based techniques (CG, SMC, TDS, and CFG). These methods are not based on RL policy optimization but are common for guiding generative models toward desired properties. This comparison validates RL-D$^{2}$ against non-RL finetuning approaches.
    \item \textbf{State-of-the-Art RL Finetuning:} We included \textbf{DRAKES}, a strong, recent baseline that also uses RL to optimize a generative model for sequence generation. DRAKES employs the Gumbel-Softmax trick to enable reward backpropagation through the generative process. This provides a direct comparison against another RL-based approach for finetuning discrete generative models.
\end{itemize}

\subsubsection{Macro Actions}
This domain tests the ability of RL-D$^{2}$ to handle online RL in complex, long-horizon tasks by modeling sequences of primitive actions.

\begin{itemize}
    \item \textbf{Standard RL Baselines:} We first included strong, general-purpose RL algorithms (DQN, IMPALA, PPO, R2D2) that operate on a single-action level. These baselines establish a performance reference and demonstrate the inherent difficulty of the tasks without temporal abstraction.
    \item \textbf{Adapted Macro-Action Baselines:} To create a direct comparison, we adapted standard algorithms to handle macro-actions, as detailed in \textbf{Appendix D.2}:
    \begin{itemize}
        \item \textbf{DQN-Macro:} This baseline represents a "naive" approach, where the Q-network's output layer is expanded to $|\mathcal{A}|^{K}$. This directly exposes the challenge of exponential scaling that RL-D$^{2}$ is designed to overcome.
        \item \textbf{IMPALA-Macro:} This baseline represents a "factored" approach, where the policy network outputs $K \times |\mathcal{A}|$ logits, assuming conditional independence between actions in the macro-action sequence.
    \end{itemize}
    These adaptations allow us to test our hypothesis that an expressive, non-causal generative model (our diffusion policy) can outperform both naive exponential-space methods and simple independent factorization methods.
\end{itemize}

\subsection{Cooperative Multi-Agent Reinforcement Learning}
This domain tests RL-D$^{2}$ on modeling the combinatorial \textit{joint action} of multiple agents.

\begin{itemize}
    \item \textbf{Autoregressive (AR) Policy:} We selected a strong \textbf{autoregressive transformer} policy as our primary baseline. This is a dominant and highly effective paradigm for modeling joint actions, where the action for each agent is sampled sequentially, conditioned on the actions of previous agents.
    \item This comparison is central to our motivation. The introduction (Section 1) explicitly notes that AR models impose an artificial causal ordering. By comparing RL-D$^{2}$ (a non-causal generative model) against a strong AR baseline, we directly test our claim that diffusion's flexible, non-causal generation process is a superior parameterization for modeling complex inter-agent dependencies in MARL.
\end{itemize}

\section{Additional Discussions.}

\textbf{Compared to discrete black-box solvers.}
Compared with standard discrete black-box solvers, our RL-D$^2$ framework is particularly well suited to sequential decision-making problems with large combinatorial action spaces. While methods such as evolutionary algorithms and zeroth-order optimization typically treat each problem instance as a static objective and often optimize from scratch, RL-$D^2$ learns a state-conditioned diffusion policy that can exploit transition dynamics, perform structured exploration, and generalize to unseen states without re-solving each instance independently. The expressive diffusion policy also provides a scalable distributional model over high-dimensional discrete actions, mitigating some of the curse-of-dimensionality issues faced by generic black-box search methods. In addition, our FKL variant supports off-policy learning, enabling reuse of previously collected trajectories or offline datasets, which can further improve query efficiency when environment interactions are expensive.